\pdfoutput=1

\documentclass[11pt]{article}

\usepackage[preprint]{acl}

\usepackage{times}
\usepackage{latexsym}

\usepackage[T1]{fontenc}

\usepackage[utf8]{inputenc}

\usepackage{microtype}

\usepackage{inconsolata}

\usepackage{graphicx}

\usepackage{xspace}                 
\usepackage{float}

\usepackage{hyperref}
\hypersetup{
    colorlinks=true,
    linkcolor=blue,
    filecolor=magenta,      
    urlcolor=blue,
    pdftitle={Overleaf Example},
    pdfpagemode=FullScreen,
    }

\usepackage{soul}
\sethlcolor{lblue}
\usepackage{makecell}
\usepackage{booktabs}
\usepackage{dsfont}
\usepackage{fdsymbol}
\usepackage{multirow}
\usepackage{colortbl}
\usepackage{longtable}

\newcommand{\hlc}[2][yellow]{{%
    \colorlet{foo}{#1}%
    \sethlcolor{foo}\hl{#2}}%
}
\colorlet{lblue}{blue!15}
\colorlet{lviolet}{violet!15}
\colorlet{lgreen}{green!35}
\colorlet{lyellow}{yellow!15}

\title{\textsc{Amulet}: Putting Complex Multi-Turn Conversations on the Stand\\ with LLM Juries
}

\author{Sahana Ramnath, Anurag Mudgil, Brihi Joshi, Skyler Hallinan, Xiang Ren  \\
Department of Computer Science, University of Southern California\\
\texttt{sramnath@usc.edu} \\
}

\newcommand{\sahana}[1]{{\color{red}{[{\bf Sahana}: #1]}}}

\newcommand{\framework}{\textsc{Amulet}\xspace}
\newcommand{\gptfour}{\textsc{GPT-4}\xspace}
\newcommand{\gpto}{\textsc{GPT-4o}\xspace}
\newcommand{\claude}{\textsc{Claude}\xspace}
\newcommand{\qwen}{\textsc{Qwen}\xspace}
\newcommand{\gptofull}{\texttt{gpt-4o-2024-08-06}\xspace}
\newcommand{\claudefull}{\texttt{claude-3-5-sonnet-20241022}\xspace}
\newcommand{\qwenfull}{\texttt{Qwen/Qwen2.5-32B-Instruct}\xspace}

\newcommand{\da}{\textsc{DA}\xspace}
\newcommand{\maxim}{\textsc{Maxim}\xspace}

\newcommand{\anthropic}{\textsc{Anthropic}\xspace}
\newcommand{\wildf}{\textsc{WildFeedback}\xspace}
\newcommand{\wildfshort}{\textsc{WildF.}\xspace}
\newcommand{\nectar}{\textsc{Nectar}\xspace}
\newcommand{\anthrtest}{\textsc{HH-Test}\xspace}
\newcommand{\anthrtrain}{\textsc{HH-Train}\xspace}

\newcommand{\io}{\textsc{I/O}\xspace}
\newcommand{\wexpl}{\textsc{W-Expl}\xspace}

\newcommand{\fjury}{\textsc{Amulet-LM-Jury}\xspace}
\newcommand{\frjury}{\textsc{Amulet-RM-Jury}\xspace}

\newcommand{\inform}{\textsc{INF-ORM}\xspace}
\newcommand{\informfull}{\texttt{INF-ORM-Llama3.1-70B}\xspace}

\newcommand{\skyllama}{\textsc{Sky-Llama}\xspace}
\newcommand{\skygemma}{\textsc{Sky-Gemma}\xspace}
\newcommand{\skyllamafull}{\texttt{Skywork-Reward-Llama-3.1-8B-v0.2}\xspace}
\newcommand{\skygemmafull}{\texttt{Skywork-Reward-Gemma-2-27B-v0.2}\xspace}
\newcommand{\armorm}{\textsc{ArmoRM}\xspace}
\newcommand{\armormfull}{\texttt{RLHFlow/ArmoRM-Llama3-8B-v0.1}\xspace}
\newcommand{\urm}{\textsc{URM}\xspace}
\newcommand{\urmfull}{\texttt{LxzGordon/URM-LLaMa-3-8B}\xspace}
\newcommand{\qrm}{\textsc{QRM}\xspace}
\newcommand{\qrmfull}{\texttt{QRM-Gemma-2-27B}\xspace}

\begin{document}
\maketitle

\begin{abstract}
Today, large language models are widely used as \textit{judges} to evaluate responses from other language models.
Hence, it is imperative to benchmark and improve these LLM-judges on real-world language model usage: a typical human-assistant conversation is lengthy, and shows significant diversity in topics, intents, and requirements across turns, e.g. social interactions, task requests, feedback. We present \textbf{\framework}, a framework that leverages pertinent linguistic concepts of \textit{dialog-acts} and \textit{maxims} to improve the accuracy of LLM-judges on preference data with \textit{complex}, \textit{multi-turn} conversational context. \framework presents valuable insights about (a) the communicative structures and intents present in the conversation (dialog acts), and (b) the satisfaction of conversational principles (maxims) by the preference responses, and uses them to make judgments. On four challenging datasets, \framework shows that (a) humans frequently ($60$-$70\%$ of the time) change their intents from one turn of the conversation to the next, and (b) in $\sim$$75\%$ of instances, the preference responses can be differentiated via dialog acts and/or maxims, reiterating the latter's significance in judging such data. \framework can be used either as a \textit{judge} by applying the framework to a single LLM, or integrated into a \textit{jury} with different LLM judges; our judges and juries show strong improvements on relevant baselines for all four datasets (code, data \href{https://github.com/INK-USC/Amulet-DA-and-Maxim}{here}).

\end{abstract}

\section{Introduction}
\begin{figure}[!t]
    \centering    \includegraphics[width=0.98\linewidth]{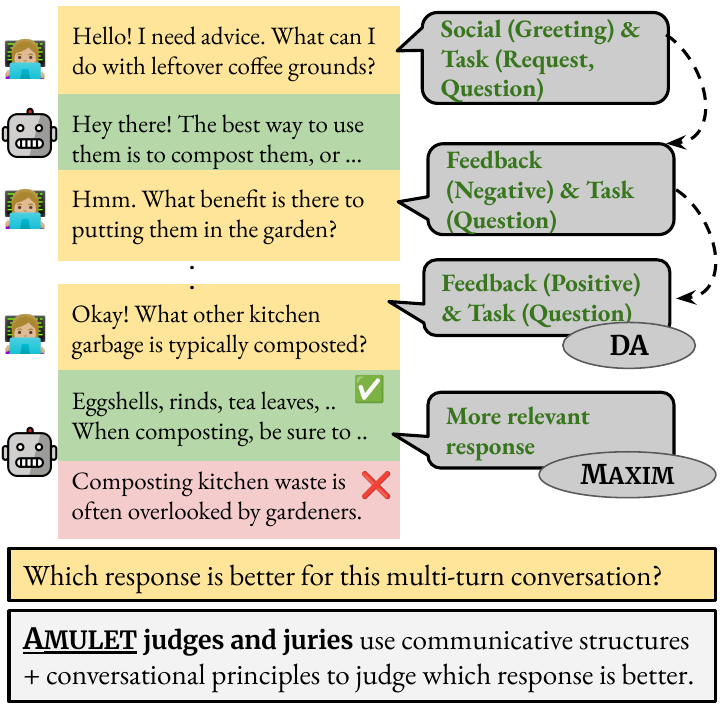}
    \caption{Real-world language model usage typically includes lengthy, complex human-assistant conversations, where humans express varying intents and requirements across the turns. Given preference data with such context, how accurate are LLM-judges in predicting which response is better? We develop a framework, \framework, that uses the following linguistic concepts for the same: (a) dialog-acts (\da) to analyze the communicative structure of each turn in the conversation, and (b) maxims to compare the preference responses in terms of principles such as informativity, truth, relevance, etc.
    }
    \label{fig:fig1}
\end{figure}



\begin{figure*}
    \centering    \includegraphics[width=0.95\textwidth]{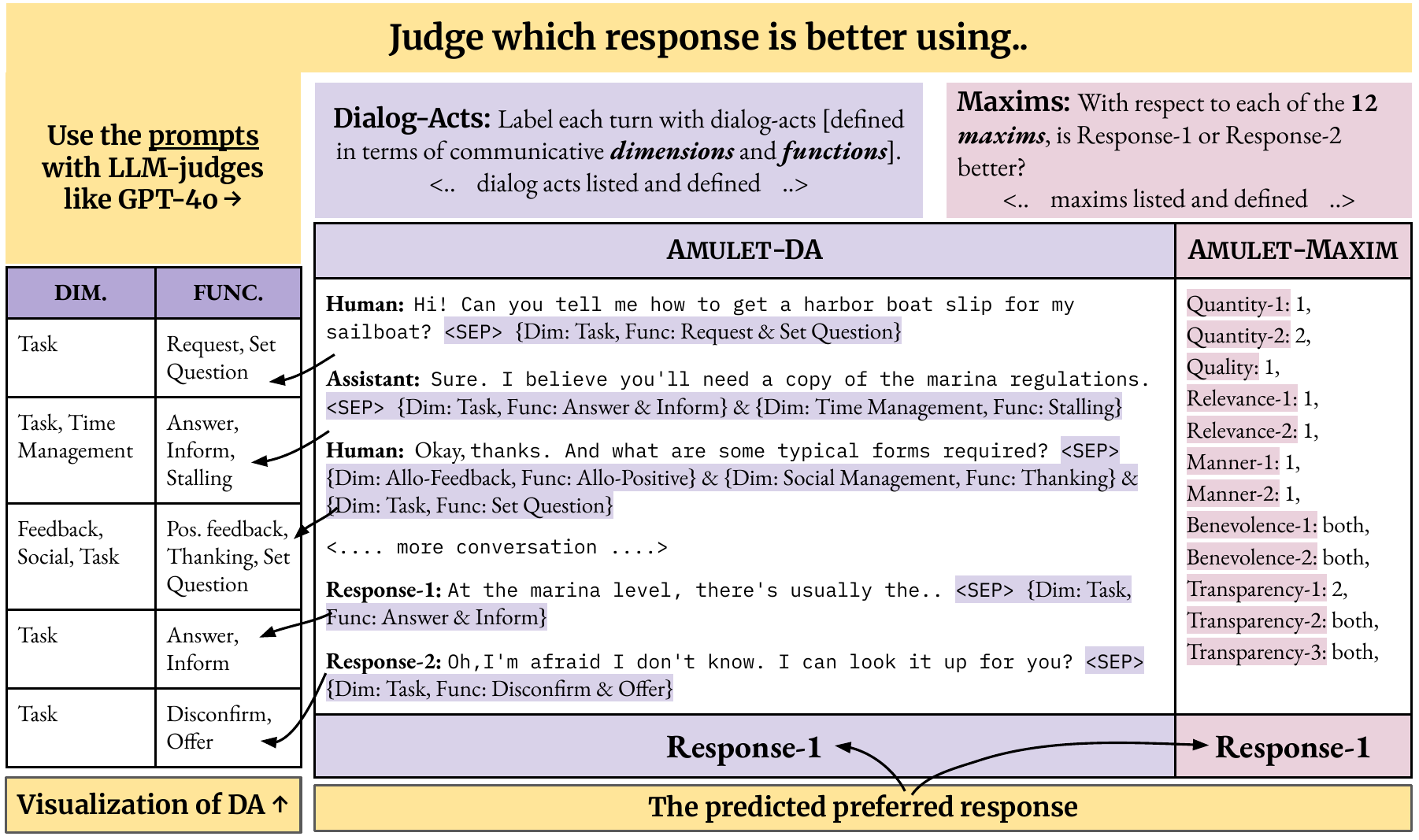}
    \caption{\framework-\da uses dialog-acts to analyze communicative structures in the conversation. In the example above, the second human turn ``Okay thanks. And what are some typical forms required?'' has the structures of feedback/positive, social/thanking and task/question. \framework-\maxim analyzes which conversational principles are satisfied by the preference responses. In the example above, Response-1 is better than Response-2 at most of the maxims. \framework uses these annotations to give more accurate preference judgments.}
    \label{fig:fig2}
\end{figure*}



In contemporary NLP literature, it has become common to prompt large language models such as \gptfour \cite{gpt4tech} as open-ended \textit{judges}, to evaluate language models \cite{mtbench,wildbench}
and to generate AI feedback to post-train them \cite{bai2022constitutional,cui2023ultrafeedback,leerlaif}.
This has made it imperative to benchmark and improve these LLM-judges on real-life usage of language model assistants by humans. 
Prior works \cite{zhao2024wildchat1mchatgptinteraction,laban2025llmslostmultiturnconversation} show that human-assistant conversations are often multi-turn, diverse, and complex in nature. However, popular LLM judge evaluation benchmarks  \cite{rewardbench,mtbench,alpaca_eval} face the following limitations: conversations in the benchmark are limited to one or two human turns, and conversations are mostly focused on individual, specific tasks (\textit{e.g.}, math reasoning, coding/debugging, logical QA, etc.) without changing requirements and intents. 
In this work, we take the first step towards analyzing and improving LLM-judges on preference datasets with complex, multi-turn conversational context (refer Figure~\ref{fig:fig1}). 

\noindent Contemporary works on LLM-judges range from generic judge prompts  ``Which response is better? Do not be biased by verbosity and position bias.'' \cite{mtbench} to  evaluation rubrics unique to each instance ``Does the response have simple vocabulary as requested in the question?'' \cite{wildbench,kim2023prometheus}; these methods are under/over-specified respectively, and not scalable to complicated, multi-turn conversations.
We overcome these issues by using linguistic theories that are pertinent to the evaluation of language models and are designed for conversational data \cite{chen2023automatic,miehling2024language}; 
we use them to develop our novel framework \textbf{\framework} (\textbf{A}nalyze \textbf{MUL}ti-\textbf{T}urn) that when applied to models such as \gptfour improves their accuracy in judging response preferences in multi-turn conversations.
Firstly, we present our novel prompts \framework-\da and \framework-\maxim (refer Figure \ref{fig:fig2}) that use dialog acts (communicative structures) and Grice's maxims (conversational principles) to accurately analyze and judge multi-turn conversational preferences.
We then present our juries \fjury and \frjury (panel of multiple judges) that integrate the above \da and \maxim judges with vanilla LLM-judges and SOTA reward models \cite{inf-orm,qrm} respectively to further improve the accuracy of judgments.

We provide experiments on four challenging datasets with 
\gpto \cite{gpt-4o} as our LLM-judge.
We observe that humans change their \da from one turn to the next with (high) probabilities of $\sim$0.7; further, the preference responses in $\sim$78\% of instances in all datasets can be differentiated via \da and/or their \maxim satisfaction numbers, indicating their significance in determining preference. We demonstrate that \framework leads to significant accuracy improvements over strong baselines, with \fjury and \frjury as our strongest methods.  
Given the wide usage of LLM-judges in both language model evaluation and synthetic preference data collection, we believe \framework can act as a light-weight add on to existing strong reward models and judges to adapt to complex human-assistant conversations.
\section{Multi-turn conversations and our framework \framework} 
\label{sec:framework}

\subsection{The complexity of multi-turn conversations} \label{sec:complexity}

Most existing preference-based evaluation systems for language models are optimized for single-turn interactions, or constrained the conversational context with a limited number of turns.
These tasks tend to focus on narrow, domain-oriented exchanges, such as solving math problems~\cite{lai2024stepdpostepwisepreferenceoptimization}, writing code~\cite{chen2021codex}, or answering factual questions~\cite{lin-etal-2022-truthfulqa}, where response evaluation is relatively localized, well-defined and based on specific rubrics, eg:- 
length, harmlessness or other predetermined axes~\cite{anthropc-hh-data}.
However, real-world human–assistant interactions are often much more complex and layered, involving longer conversational threads, shifts in user intent, social dynamics, and varying types of information needs over time (Appendix \ref{app:example-complex-mt}). As such, current benchmarks fall short in capturing the nuances required to evaluate assistant responses in extended, evolving conversations.
To this end, selecting the right benchmark of multi-turn conversations 
is extremely challenging.
To select appropriate datasets, we utilized the following heuristics.
First, the dataset should have multi-turn dialogues between the human and the AI assistant, with preference choices for the last assistant response.
Second, given wide contamination of preference benchmarks with reward models~\cite{rewardbench}, we aim to choose benchmarks that have the least or lowest contamination with the models that we select to evaluate, to provide a fairer evaluation.
Lastly, several preference benchmarks are often synthetic in nature ~\cite{dong2024selfboostinglargelanguagemodels,singh2025fspofewshotpreferenceoptimization}.
We aim to select benchmarks that include realistic human-assistant conversations \cite{zhao2024wildchat1mchatgptinteraction}.

\paragraph{Benchmarks for experiments.} We provide experiments on three datasets: \anthropic (`helpful' subset), \wildf and \nectar. 
We select instances wherein the conversations have $\geq4$ human turns; we also provide results on the $\geq 7$ human turns subset of the above. 
For \anthropic we report results on both train (\anthrtrain) and test (\anthrtest) splits.
For \anthrtest we do not provide results on the $\geq 7$ turns subset since it does not have enough instances with the same.  Note that these datasets already have preference labels, i.e., the responses have been classified into the chosen and the rejected (there are no instances with the label of `tie').  
In Appendix \ref{app:data-cleaning} we detail our data cleaning strategies and final dataset sizes.

\paragraph{Terminology.} In this work, a human-assistant conversation consists of alternating turns of conversation between a \textbf{human} and a language model \textbf{assistant}, always initiated by the human and always ended by the assistant. 
In prior work \cite{mtbench}, a ``turn'' could jointly refer to a human query and the immediately following assistant response; in this work, we use the term ``turn'' to refer to either the human's or the assistant's utterance. 
We define the length of a conversation by the \textit{number of human-turns} it contains. 
We use the term (evaluation) `\textbf{instance}' to refer to an entire multi-turn conversation with alternating human and assistant turns, with the last assistant turn having two \textit{preference} responses. 
Given an instance $E$ with preference responses $R_1$ and $R_2$, we want our judge $M$ to predict the preferred response $R_{pref} \in \{R_1, R_2\}$.

\subsection{Dialog Acts: \framework-\da}
We analyze the communicative structure of  conversations via \textit{dialog acts} \cite{allen1997draft,bunt2011multifunctionality,chen2023automatic}. 
Prior work discusses how turns in conversations are \textit{multi-functional} in nature, that is, they can serve more than one communicative function. \citet{bunt2011multifunctionality} formalizes this concept by defining two terms: communicative \textit{dimensions} that deal with the semantic/information content, and communicative \textit{functions} that deal with the linguistic phrasings; they propose an ISO standard annotation scheme for the same in \citet{bunt2019guidelines}, which we present in Table \ref{tab:dialog-acts}.
We design prompt \textbf{\framework-\da} (Appendix \ref{app:da-prompt}) that when used with a judge $M$, enables the judge to predict the dimensions and functions $E_{DA}$ present in each turn of the instance $E$'s conversation and to use these dialog act predictions to pick the preferred response $R_{pref}$. 

\begin{equation}
    R_{pref}, E_{DA} = M(E, \text{\framework-\da})
\end{equation}




\begin{figure*}
    \centering    \includegraphics[width=0.98\textwidth]{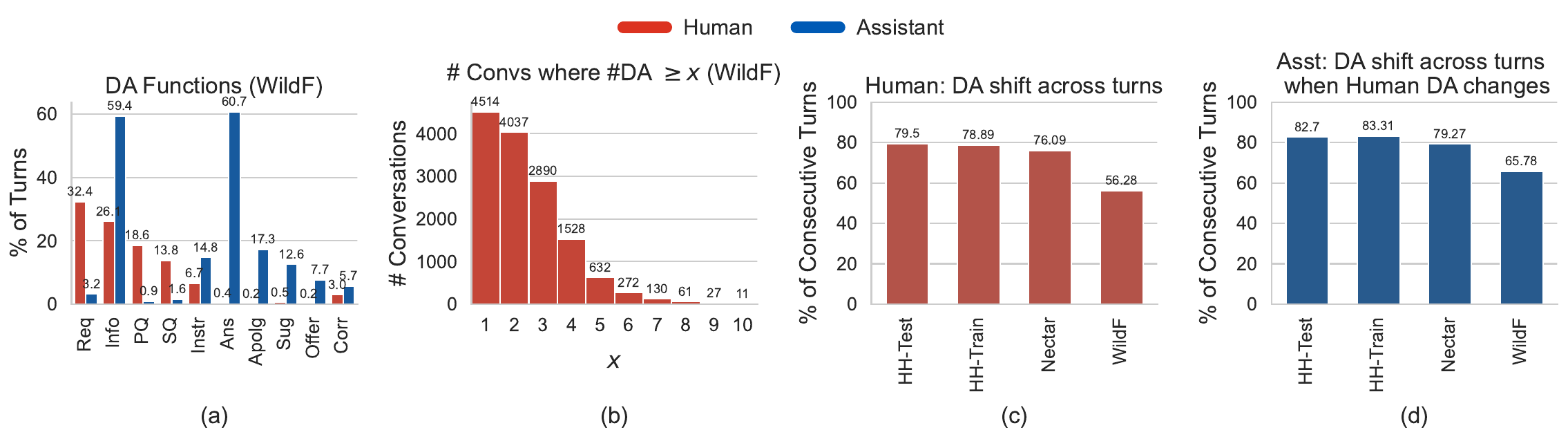}
    \caption{\textbf{Graphs for Section \ref{sec:analyze-conv}:} (a) Frequency of most common functions in \wildf, (b) Distribution of \#conversations where the human turns' $\#\da's \geq x$ for \wildf, (c) \% of consecutive human turns with different \da's, (d) \%  of consecutive assistant turns with different \da's \textit{when} the corresponding consecutive human \da's are different. (acronyms, eg:- PQ is Propositional Question, SQ is Set Question, etc. in Appendix \ref{app:remaining-analysis-conv-figs})}
    \label{fig:da-distribution}
\end{figure*}

\subsection{Grice's Maxims: \framework-\maxim}
We use the fundamental conversational principles of \textit{Gricean Maxims} \cite{grice1975logic} to analyze 
language model responses in terms of how they (don't) satisfy conversational principles.
Grice proposed four fundamental maxims of conversation, \textit{Quality} (truth), \textit{Quantity} (informativity), \textit{Manner} (clarity), and \textit{Relation} (relevance) that are followed in a conversation to have an effective and cooperative interaction. 
Recently, \citet{miehling2024language} extended Grice's maxims to human-AI interactions.
They broke down the original four maxims into multiple sub-parts that have clear definitions ; for example, Quantity became Quantity-1 (the response should have enough information) and Quantity-2 (the response should not have unnecessary information). They also introduced two new maxims specific to AI assistants, Benevolence (moral responsibility) and Transparency (recognizing constraints and boundaries). Table \ref{tab:maxims} in Appendix \ref{app:da-and-maxim} demonstrates all the maxims and sub-maxims with their definitions. We adopt the twelve sub-maxims proposed by \citet{miehling2024language} in our work (for ease of reference, we refer to the sub-maxims as maxims in the rest of the paper). We design prompt \textbf{\framework-\maxim} (Appendix \ref{app:maxim-prompt}) that when used with a judge $M$ enables to judge to predict which of the two preference responses $R_1,R_2$ satisfy each maxim better $E_{Maxim}$ and use this analysis to pick the preferred response $R_{pref}$.
\begin{equation}
    R_{pref}, E_{Maxim} = M(E, \text{\framework-\maxim})
\end{equation}

\section{Analyzing multi-turn conversations and preferences with \framework} 
\label{sec:analysis}



\subsection{What is in a multi-turn conversation?}
\label{sec:analyze-conv}
We analyze patterns of dialog acts in multi-turn conversations to understand the latter's challenging nature. We apply \framework-\da to each conversation\footnote{For this section, we consider the conversation until the last human turn.}, extract the dialog acts for each turn and analyze them in terms of frequency and transitions across turns. Note that a turn's dialog act could comprise of multiple dimensions and functions; when we say a turn has the dialog act of $\da_i$, we are referring to the combination of all dimensions and functions present in the turn. For example, if a speaker says ``Okay, thanks!'', the dialog act is ``Dimension: Allo-Feedback with Function: Positive Feedback'' + ``Dimension: Social with Function: Thanking'' (full example in Appendix \ref{app:annotation}).
 .



\begin{table*}[ht]
\centering
\resizebox{0.98\textwidth}{!}{%
\begin{tabular}{lcccccccc}
\toprule
\multirow{2}{*}{\textbf{\#\maxim's satisfied}} 
& \multicolumn{2}{c}{\textbf{\anthrtest}} 
& \multicolumn{2}{c}{\textbf{\anthrtrain} } 
& \multicolumn{2}{c}{\textbf{\nectar}} 
& \multicolumn{2}{c}{\textbf{\wildf}} \\
\cmidrule{2-9}
& Same \da & Diff. \da 
& Same \da & Diff. \da 
& Same \da & Diff. \da 
& Same \da & Diff. \da \\
\midrule
Chosen satisfies more   & 15.9\%   & 59.1\%   
& 16.3\% & 59.1\% & 28.5\% & 52.1\%  & 51.3\% & 30.3\% \\
Rejected satisfies more & 4.3\% & 15.3\%  
& 4.2\% & 14.4\%  & 2.9\%   & 3.3\% & 4.7\% & 4.3\% \\
Both satisfy equally & 2.6\% & 2.8\%  & 2.8\%  & 3.2\% & 9.3\%   & 3.9\% & 7.6\% & 1.6\%    \\ \midrule
\textbf{Total}                 & \textbf{22.8\%} & \textbf{77.2\%} 
& \textbf{23.3\%} & \textbf{76.7\%} 
& \textbf{40.7\%} & \textbf{59.3\%} 
& \textbf{63.7\%} & \textbf{36.3\%} \\
\bottomrule
\end{tabular}
}
\caption{Distribution of Grice’s maxim satisfaction across datasets split by Same vs. Different Dialog Acts (DAs). All \% values are measured on the size of the dataset, for example, 15.9\% of instances \anthrtest have chosen/rejected samples have the same \da, with the chosen satisfying more \#\textsc{Maxim}s.}
\label{tab:maxim_summary_all}
\end{table*}

\begin{figure*}
    \centering
    \includegraphics[width=0.98\textwidth]{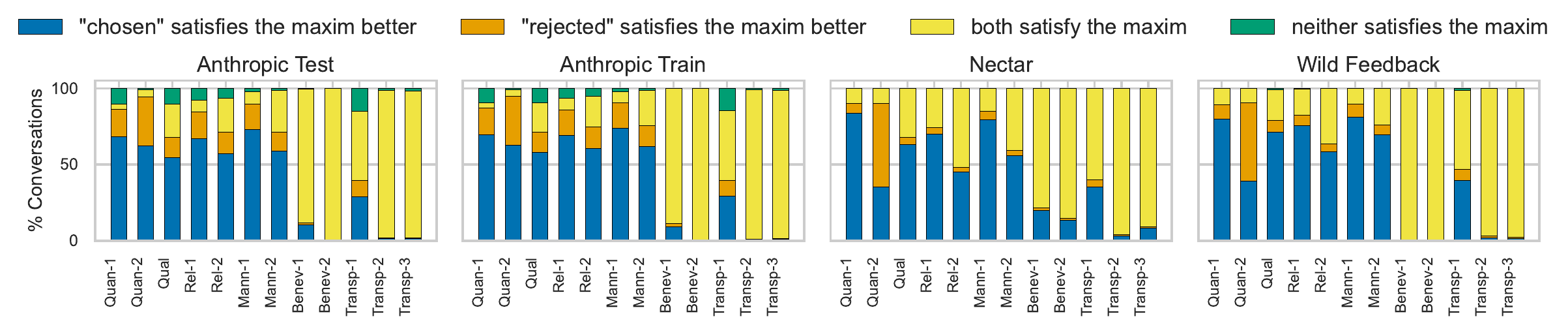}
    \caption{For each maxim on the x-axis, we measure the \% of conversations in the dataset where (a) the chosen response satisfies the maxim better than the rejected, (b) the rejected response satisfies the maxim better, (c) both responses satisfy the maxim equally, (d) neither response satisfies the maxim. We see that maxims such as Quantity-1, Relevance-1, etc. are important in all datasets to distinguish between the chosen and rejected responses, and maxims such as Benevolence-2 are only significant in certain datasets, \nectar.}
    \label{fig:maxim-importance}
\end{figure*}

\paragraph{Both human and assistant turns are found to have a varied set of communicative functions (Figures \ref{fig:da-distribution}(a), \ref{fig:freq-funcs-other}, \ref{fig:freq-dims}).} We analyze the frequency distribution of dialog acts in human and assistant turns; we find that for all four datasets, the most common dimension for both humans and assistants is `Task' (regarding the underlying task), followed by `Social Obligations Management' (regarding social interactions such as greeting, thanking) or `Allo-Feedback' (regarding the addressee's processing of the previous turn) (Figure \ref{fig:freq-dims}). The most common functions vary slightly across datasets (Figures \ref{fig:da-distribution}(a), \ref{fig:freq-funcs-other}).  In general, human turns show a high proportion of information seeking functions such as questions and requests,  social functions such as thanking, and giving feedback, and assistant turns show a high proportion of information providing functions such as inform, offer, and suggest, and social functions such as apology. 



\paragraph{More than half the instances have \textit{three or more} dialog acts in the conversation (Figures \ref{fig:da-distribution}(b), \ref{fig:cdf-other}).} We aim to find if conversations are monolithic with respect to dialog acts, or if they show a large diversity. For $x \in \{1,2,3,..\}$, we count the number of conversations where the number of dialog acts is equal to or more than $x$ (i.e., $\#\da's \geq x$). We present the results as graphs (Figures \ref{fig:da-distribution} (b) and \ref{fig:cdf-other}). We observe that almost all instances have atleast 2 dialog acts in their human turns, and more than half the instances have atleast 3 dialog acts.   


\paragraph{Humans change their dialog acts from one turn to the next $\sim$$73\%$ of the time, and in response, assistants change their dialog acts $\sim$$78\%$ of the time (Figure \ref{fig:da-distribution} (c,d)).}
We hypothesize that assistants often mirror the dynamic nature of human multi-turn conversations, which necessitates our framework to appropriately characterize both human and assistant turns.
We calculate the proportion of \textit{consecutive} human turns which show a change in \da. 
This serves to show that humans change the type of their requests frequently, subsequently requiring assistants to adapt and satisfy their varying demands. 
For all datasets, we show the \% of consecutive turns with different \da in Figure \ref{fig:da-distribution} (c). Next, for the consecutive human turns with \da changes, we calculate the \% of corresponding assistant responses that also show a change in \da; this serves to show how often assistants modify their \da's in response to the human's changing intents. For all datasets, we show the \% of consecutive assistant turns with different \da when their corresponding human turns also show a change in \da in Figure \ref{fig:da-distribution} (d).


\subsection{Preference responses differ significantly  with respect to \da and \maxim} \label{sec:analyze-pref}
In this section, we compare the chosen and rejected responses in terms of their dialog acts, and the number of maxims they respectively satisfy. 
\paragraph{Dialog Acts and Maxims provide signals to distinguish between response preferences in $\sim$$70$-$80\%$ of instances (Table \ref{tab:maxim_summary_all}).} Are dialog acts and maxims strong signals to distinguish between preferences? We first obtain the \% of the instances in each dataset where the preference responses have the same or different \da's. 
Then, within these categories, we obtain \% of instances where the chosen response satisfies more, less or the same \#\maxim when compared to the rejected response.
We see that for \anthrtrain, \anthrtest and \nectar, most instances' preference responses have different dialog acts ($77\%, 77\%, 60\%$), although for \wildf, a majority of the instances have the same \da ($64\%$). Further, we see that the chosen response satisfies more maxims than the rejected response in $\sim$$80\%$ of the instances within both categories.

\paragraph{Maxims Quantity, Quality, Relevance and Manner are significant to distinguish preference in all datasets (Figure \ref{fig:maxim-importance}).} Finally, we analyze which maxims are the most important to distinguish preferences. 
For each maxim, we measure the \% of instances in each dataset for which: the chosen response satisfies the maxim better, the rejected response satisfies the maxim better, both responses satisfy the maxim, and neither response satisfy the maxim. We see that in all datasets, there are a significant \% of instances where the chosen response satisfies the maxims of Quantity, Quality, Relevance, Manner and Transparency-1 better; this indicates the importance of these maxims in picking the better response.
Benevolence-1 is marginally important for three out of the four datasets. For \nectar, all the maxims seem to be important at various levels; for the other three datasets, maxims like Transparency-3 almost play no role in distinguishing the chosen from the rejected, since both seem to satisfy it equally.

\section{Experiments: Judge and Jury} \label{sec:exps}

\subsection{Experimental Setup} \label{sec:exp-setup}

\paragraph{Evaluating LLM Judges.}
Given a conversation and two preference responses, we measure the \textbf{\textit{accuracy}} of the judge/jury picking the better (chosen) response (similar to  \citet{rewardbench}). We use \gpto (\gptofull) with a temperature of $0$ in all our prompt-based experiments.

\paragraph{Addressing position bias.} As seen in our prompts \framework-\da and \framework-\maxim, we give the models the prompt, followed by the conversation, followed by the preference responses. However, prior works \cite{mtbench} have noted that models exhibit \textit{position bias}, that is, they tend to pick the first presented option as the answer. To prevent this, we use refer to prior work \cite{mtbench} and follow the simple solution of running each instance twice, with the preference responses having swapped positions both times\footnote{The first time, the responses are shown in the order $R_1, R_2$. The second time, the responses are shown in the order $R_2, R_1$.}. Hence, we get two predicted answers per instance; we term these as \textit{votes}. We consider that the model has judged the instance correctly if and only if it picks the chosen response in each vote.


We note that the models we experiment on \textit{may} have seen \anthrtrain in their training (we do not know for sure, since these models do not disclose the data they were trained on); however, as we see below in our results, the trends we observe in our results hold for \anthrtrain as well. 



\subsection{Comparison Models}
\paragraph{Standard LLM-judge.} We compare with two LLM-judge baselines, \io and \wexpl with \gpto (prompts in Tables \ref{tab:io-prompt}, \ref{tab:wexpl-prompt}). 
Given the prompt and the evaluation instance, the \io judge outputs which response is better ($R_{pref} \in \{R_1,R_2\}$); the \wexpl judge outputs $R_{pref}$, as well as a natural language explanation for the same.
We design these prompts based on prior works such as \citet{mtbench,lee2024aligning}. We use the two-vote system we proposed in Section \ref{sec:exp-setup}.

\paragraph{SOTA Reward Models (RMs).} We also compare with three state-of-the-art reward models of varying architectures and sizes taken from the top of RewardBench leaderboard \cite{rewardbench}: \inform \cite{inf-orm}, \qrm \cite{qrm} and \skyllama \cite{skywork}\footnote{\informfull, \qrmfull, \skyllamafull}. These models have been specifically developed to judge the quality of language models' responses via a combination of high-quality training data, targeted objectives and sometimes multiple stages of training (details in Appendix \ref{app:sota-rms}); these make the RMs extremely strong competitors to our methods. All three RMs are score-based, i.e., they take in a conversation and \textit{one} response and provide an unbounded numerical score. The response with the higher score is selected as the RM's preferred response. In the unlikely event of the RM producing the exact same score for both preference responses, we deem the RM to have failed at that instance. 
Since these models take in one response at a time, they are \textit{not prone} to position bias. Note that we chose RMs which have not been trained on any of the datasets we use for evaluation.

\begin{table*}[h]
\centering
\resizebox{0.98\textwidth}{!}
{%
\begin{tabular}{l|cccc|ccc}
\toprule
Dataset $\rightarrow$ & \anthrtest & \wildfshort & \nectar & \anthrtrain & \wildfshort & \nectar & \anthrtrain \\ 
Method $\downarrow$, $\#$ human turns $\rightarrow$ & \multicolumn{4}{c|}{$\geq 4$} & \multicolumn{3}{c}{$\geq 7$} \\ \midrule
\io (\gpto) & 55.7  & 75.1  & 75.8  & 57.6 & 75.0 & 79.7  & 53.3 \\ 
\wexpl (\gpto) & 55.9 & 75.0 & 75.9 & 57.8 & 74.5 &  81.8 & 55.3 \\ \midrule 
\skyllama & 66.3 & 78.7 & 72.1 & 65.8 & 78.8 & 70.7 & 65.7 \\ 
\qrm & 63.3 & 72.3 & 75.0 & 64.8 & 72.6 & 74.9 & 63.3 \\ 
\inform & 68.7 & 84.1 & 77.7 & 67.4 & 84.3 & 77.6 & 66.1 \\ \midrule \midrule
\framework-\da (\gpto) &  \cellcolor{lyellow}  59.8 & \cellcolor{lyellow}  72.7 & \cellcolor{lyellow}  76.9 & \cellcolor{lyellow}  60.2 &  \cellcolor{lyellow} 72.8 &  \cellcolor{lyellow} 79.1  &  \cellcolor{lyellow} 58.8 \\
\framework-\maxim (\gpto) &  \cellcolor{lyellow} 49.8 & \cellcolor{lyellow} 62.2 &  \cellcolor{lyellow} 69.9 &  \cellcolor{lyellow} 50.9 &  \cellcolor{lyellow} 60.6 &  \cellcolor{lyellow} 76.7  &  \cellcolor{lyellow} 45.8 \\ \midrule
\da-then-\maxim (\gpto) & \cellcolor{lviolet} 62.6 & \cellcolor{lviolet}  76.9 & \cellcolor{lviolet}  79.1 & \cellcolor{lviolet}  63.5 & \cellcolor{lviolet}  76.3 & \cellcolor{lviolet}  83.6  & \cellcolor{lviolet}  61.1 \\
\da-then-\maxim-then-\wexpl (\gpto) & \cellcolor{lviolet}  64.1 &  \cellcolor{lviolet} 80.2 &  \cellcolor{lviolet} 80.5 &  \cellcolor{lviolet} 66.2 &  \cellcolor{lviolet} 79.8 & \cellcolor{lviolet}  86.0 &  \cellcolor{lviolet} 63.3 \\ \midrule
\da-then-\maxim-then-\skyllama & \cellcolor{lblue}  67.0 & \cellcolor{lblue}  83.6 & \cellcolor{lblue} 82.8 & \cellcolor{lblue}  68.6 & \cellcolor{lblue}  84.1 & \cellcolor{lblue}  86.3 &  \cellcolor{lblue} 66.6 \\
\da-then-\maxim-then-\qrm & \cellcolor{lblue}  67.0 & \cellcolor{lblue}  82.6 & \cellcolor{lblue}  82.6 & \cellcolor{lblue}  68.8 & \cellcolor{lblue}  83.3 & \cellcolor{lblue}  85.7 & \cellcolor{lblue}  66.2 \\
\da-then-\maxim-then-\inform & \cellcolor{lblue} 67.0 &  \cellcolor{lblue} 84.2 &  \cellcolor{lblue} 83.1 & \cellcolor{lblue} 68.7 &  \cellcolor{lblue} 85.0 & \cellcolor{lblue} 86.9 & \cellcolor{lblue} 67.0 \\ \bottomrule
\end{tabular}%
}
\caption{\textbf{All experimental results.} \fjury and \frjury show significant accuracy improvements over vanilla LLM-judges and SOTA reward models. Depending upon the available compute, either one of these juries can be used to improve judgment accuracies for complex multi-turn conversations.
}
\label{tab:all-exps}
\end{table*}

\subsection{Our Methods}
\paragraph{\framework.} 
We use \framework-\da and \framework-\maxim with the aforementioned two-vote system.

\paragraph{\fjury.} 
\begin{figure}[!t]
    \centering    \includegraphics[width=0.85\linewidth]{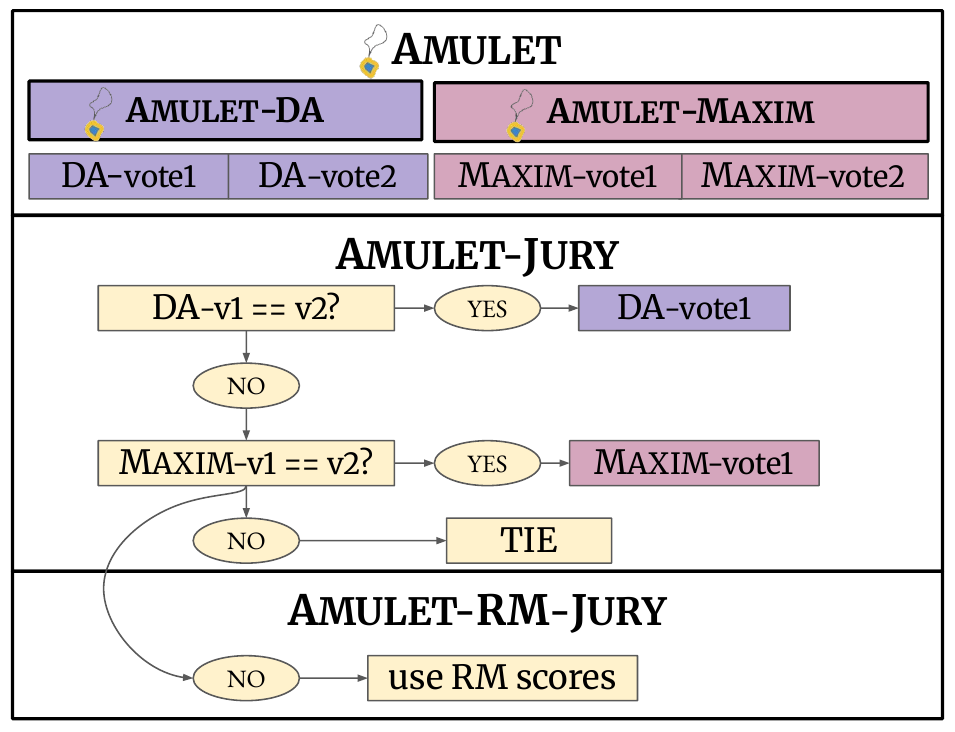}
    \caption{Voting pipeline in \fjury and \frjury.}
    \label{fig:fig3}
\end{figure}

The \da and \maxim methods use different principles to derive the final answer.
To identify the correct follow-up response to a conversation, a judge could need to use \textit{either} or \textit{both} of these principles. Hence, we propose to make an \fjury by leveraging the two-vote system we use for \framework-judges. Our first \fjury uses \framework-\da and \framework-\maxim, and we call it \textbf{\da-then-\maxim} (refer Figure \ref{fig:fig3}). We first consider the votes given by \framework-\da; if they agree on a response (i.e., both point to the same response), we use that as this jury's final choice; however, if they \textit{disagree} (i.e., it's a tie), we move on to \framework-\maxim. If \framework-\maxim's votes agree on a response, we now use that as this jury's final choice; if these votes \textit{also disagree} on which response to pick, the jury has failed on this instance.

Our second \fjury extends the above with the \wexpl judge. In the \da-then-\maxim jury, the latter method is used to break the tie (if any) in the former method. However, this jury can still result in a tie if both the \da and the \maxim methods individually resulted in a tie. We hypothesize that such cases happen when the model is unable to use either of these concepts to conclusively pick an answer; to deal with such cases, we extend the jury to include the implicit logic of the vanilla \wexpl judge. In this vein, we propose the jury \textbf{\da-then-\maxim-then-\wexpl}. 

\paragraph{\frjury.} 
We propose \frjury to leverage the strengths of both \framework and the RMs. We extend the \da-then-\maxim jury to \textbf{\textsc{\da-then-\maxim-then-RM}}; whenever \da-then-\maxim ends in a tie, we use the scores provided by the RM at hand to make the final decision (refer Figure \ref{fig:fig3}).



\subsection{Results}
In Table \ref{tab:all-exps}, we present the accuracies obtained by all judges, RMs and juries for the four datasets. 

\paragraph{\framework.} Firstly, in almost all datasets/settings, the \framework-\da judge outperforms the baseline LLM-judges \io and \wexpl. \framework-\maxim has lower accuracies, leading to the observation that maxims on their own are insufficient to predict the correct response preference in a conversation; however, we hypothesize that maxims still catch valuable distinctions between the preference responses which are missed by dialog acts. We verify this with \fjury.

\paragraph{\fjury.} When we combine dialog acts and maxims as a jury in \da-then-\maxim, we see a notable increase in accuracies across the board, confirming our prior hypothesis. The next \da-then-\maxim-then-\wexpl leads to further improvements, and is the best performing LLM-only jury; this jury beats \skyllama (size 8B) and \qrm (size 27B) in almost all cases but falls short of the largest reward model \inform (size 70B). We note here that if there is a need for a high-quality judgments for multi-turn preference data, by researchers/users who do not have access to the compute resources necessary to run fine-tuned RMs, \da-then-\maxim-then-\wexpl is a competitive alternative for complex multi-turn inputs. 

\paragraph{\frjury.} Finally, we see that all \frjury's lead to even better accuracies across the board improving over both \framework and strong, fine-tuned RMs. There are significant improvements in accuracy in all cases over \skyllama and \qrm; apart from the cases of \anthrtest and \wildf ($\geq4$ turns), \frjury also heavily improves over \inform. We conclude that a researcher/user who has access to both strong LLM-judges such as \gpto and to compute resources for RMs can use our \frjury to obtain the best judgments.

Hence, we see that \framework is a convenient and light-weight supplement to strong LLM-judges as well as SOTA RMs, that helps to significantly improve accuracy of preference judgments.

\subsection{More analyses}
\paragraph{Varying model architectures.}
We use \framework with \claude \cite{claude3} and \qwen\footnote{\claudefull, \qwenfull}\cite{qwen2.5} on \anthrtest (Table \ref{tab:anthropc-text-exps}), to establish the efficacy of our framework across LLM-architectures; we observe similar trends as in Table \ref{tab:all-exps}, with \da-then-\maxim-then-\skyllama with \qwen leading to the highest accuracy for \anthrtest, beating \inform.
\begin{table}[h]
\centering
\resizebox{0.98\linewidth}{!}{%
\begin{tabular}{l|ccc}
\toprule
Method $\downarrow$, Model $\rightarrow$ & \gpto & \claude & \qwen \\ \midrule
\io & 55.7 & 59.3 & 54.8 \\
\wexpl & 55.9 & 59.1 & 55.0 \\ \midrule \midrule
\framework-\da & \cellcolor{lyellow} 59.8 & \cellcolor{lyellow} 59.8 & \cellcolor{lyellow} 51.1 \\
\framework-\maxim & \cellcolor{lyellow} 49.8 & \cellcolor{lyellow} 59.8 & \cellcolor{lyellow} 46.5 \\ \midrule
\da-then-\maxim & \cellcolor{lviolet} 62.6 & \cellcolor{lviolet} 65.9 & \cellcolor{lviolet} 61.3 \\
\da-then-\maxim-then-\wexpl & \cellcolor{lviolet} 64.1 & \cellcolor{lviolet} 68.0 & \cellcolor{lviolet} 65.9 \\ \midrule
\da-then-\maxim-then-\skyllama & \cellcolor{lblue}  67.0 & \cellcolor{lblue}  67.8 & \cellcolor{lblue}  \textbf{69.6} \\ 
\da-then-\maxim-\qrm & \cellcolor{lblue}  67.0 & \cellcolor{lblue}  67.6 & \cellcolor{lblue}  69.3 \\
\da-then-\maxim-then-\inform & \cellcolor{lblue}  67.0 & \cellcolor{lblue}  67.0 & \cellcolor{lblue}  68.5 \\ \bottomrule
\end{tabular}%
}
\caption{\anthrtest results across model variants.}
\label{tab:anthropc-text-exps}
\end{table}

\begin{table}[h!]
\centering
\resizebox{0.98\linewidth}{!}{%
\begin{tabular}{l|ccc}
\toprule
Dataset & \multirow{2}{*}{\da} & \multirow{2}{*}{\da-then-\maxim} & \da-then-\maxim \\  
 (win / tie / loss) & & & -then-\wexpl \\ \midrule
\anthrtest & 59.8 / 15.9 / 24.3 & 62.6 / 9.1 / 28.3 & 64.1 / 5.0 / 30.9 \\ \midrule
\wildfshort & 73.0 / 16.3 / 10.6 & 76.9 / 10.4 / 12.7 & 80.2 / 5.8 / 14.0 \\ \midrule
\nectar & 76.9 / 10.5 / 12.6 & 79.1 / 7.3 / 13.5 & 80.5 / 5.1 / 14.4 \\ \midrule
\anthrtrain & 60.3 / 15.1 / 24.6 & 63.5 / 9.5 / 27.0 & 66.2 / 4.7 / 29.1 \\ \bottomrule
\end{tabular}%
}
\caption{Win-Tie-Loss statistics observed for an \framework-judge (\da) and two \fjury's. The juries have higher win and loss rates  and a lower tie rate as compared to the judge \da, proving that our juries help to \textit{improve accuracy} and \textit{break the tie}.}
\label{tab:win-tie-loss}
\end{table}
\paragraph{Win vs. Tie vs. Lose} \label{sec:win-tie-lose} Since our methods produce two votes for their final answer, their end result could be a \textit{win} (both votes lead to the correct better response), a \textit{tie} (the votes disagree on the response to be picked), or a \textit{loss} (the votes agree on the wrong response). In this section, we analyze the win/tie/loss statistics of some methods from Section \ref{sec:exps} in Table \ref{tab:win-tie-loss}; we analyze the \framework-\da judge and two \fjury's.
We see that the juries have a higher win rate (as evidenced by the higher accuracies) than the judge; the juries also have a lower tie rate and a higher loss rate as compared to the judge, indicating the helpfulness of our jury to break more ties. 

\paragraph{Qualitative Analysis.} We qualitatively investigate some instances that resulted in either a tie or a loss in Appendix \ref{app:qualitative-analysis}. We find a notable number of samples in the datasets where the preference responses are either semantically equivalent or both equally plausible; instances like these naturally lead to a tie. Hence, we stress the need for higher quality preference data collection which either removes such instances from the dataset, or allows a third label of `tie'. We also show examples where \framework-\da's tie was broken by \framework-\maxim; we observe two trends: (1) if the dialog acts of the two responses are almost exactly the same, maxims can help find which is the better response, and (2) if the dialog acts of the two responses are different but still similarly plausible, maxims play a large role in determining which response is more preferred. 







\section{Conclusion}
We develop and present \textbf{\framework}, our framework that uses linguistic concepts of dialog acts and maxims to provide more accurate judgments for multi-turn conversational preference data. \framework's judges and juries leads to notable improvements in accuracies on four challenging datasets; \framework proved to be a successful and easy supplement to strong SOTA reward models and LLM-judges to improve accuracy. In the future, we hope to improve on this further with more intricate linguistic features such as dependence relations and qualifiers \cite{bunt2019guidelines}; we also hope to extend \framework to fine-tuned, smaller language models, for more economical and accessible deployment. \textbf{Note:} Code and datasets are uploaded at \href{https://github.com/INK-USC/Amulet-DA-and-Maxim}{\texttt{github.com/INK-USC/Amulet-DA-and-Maxim}}.

\section*{Limitations} \label{sec:lim}
In this section, we provide relevant limitations, as well as the steps we take to mitigate them. 

\paragraph{Data contamination.} Owing to the large amount of training data used for pre-training, fine-tuning and aligning large language models (which are mostly not open sourced), data contamination is a massive problem in NLP today \cite{sainz2023nlp}. Most relevantly, several reward models submitted to the RewardBench leaderboard \cite{rewardbench} were contaminated with the evaluation data which leads to the observation that this is a major issue when it comes evaluating reward models and judges today. To the best of our knowledge, all the models (including SOTA reward models) used in this work are free of data contamination with respect to our evaluation data. As we mention in Section \ref{sec:exp-setup}, we do not know with certainty if the judges and RMs we use have been exposed to \anthrtrain since our judges \gpto, \claude, \qwen and our reward models  (finetuned on top of) \textsc{Llama-8/70B}, \textsc{Gemma-27B} do not release details regarding their training. We acknowledge this before presenting our results; moreover, we note that trends on \anthrtrain are similar to the trends observed on other datasets, which confirms that even if the model had seen \anthrtrain during training, that does not affect our results.



\paragraph{Evaluation dataset and method limitations.} We encountered a number of instances in the datasets wherein the preference responses were either highly similar or both equally plausible (Appendix \ref{app:qualitative-analysis}); it is unfair to have such instances in our evaluation, since we are essentially expecting the judge/jury to pick the response the original dataset has picked as the chosen response (based on their annotators' individual criteria or randomly), when both responses are equally probable. 
While we do not know that this might be an individual preference, existing LLM judges are not equipped to model an individual user's preferences~\cite{dong2024llmpersonalizedjudge}.
This is a systematic dataset problem; today's preference datasets rarely allow the label of \textit{tie}. We stress that preference datasets should either remove such instances, or if not, allow their annotators to mark the responses as tied. Lastly, we note that if the multi-turn conversation or the preference responses are too task specific (eg:- the entire conversation is just editing and debugging code) or simulated \cite{capybara}, dialog acts and maxims cannot be used to successfully distinguish between the responses. 

\paragraph{Practical limitations.} Some instances with lengthy conversations and/or preference responses fail when running with \qwen or the reward models due to GPU memory issues. We err on the side of caution, and consider such situations to be a loss in our accuracy calculation. Lastly, owing to the high costs of using \gpto and \claude (Appendix \ref{app:reproducibility}), we only report single runs of experiments; to verify the stability of our method we rerun \framework-\da with \anthrtest on \gpto. We find that the second run results in an accuracy of yields an accuracy of 58\%, almost equivalent to the accuracy obtained in the original run.   

\paragraph{Hallucination and variation in output generation.} Language models are prone to hallucinate details that are not factual or are not present in the provided context \cite{ji2023survey,rawte2023troubling,huang2023survey}; in this work, it is possible for the judges to hallucinate dialog acts and/or details in the conversation when performing the task. We report statistics of hallucination of dialog acts in Appendix \ref{app:valid-stats}; since we rigorously stress in our prompts in Appendix \ref{app:da-prompt}, \ref{app:maxim-prompt} that the judge \textit{should not make up new dialog acts}, we observe that the hallucination values in predicting dialog are extremely low $\sim$$0.1\%$. However, the model could still hallucinate details that are not present in the dialog when generating the natural language explanations while using the \da, \maxim and \wexpl prompts; while these do not affect the final accuracy of the method, they still pose a limitation with respect to the interpretability of the model and the user/researcher's trust in the model when using it for downstream tasks. Contemporary works work on analyzing and mitigating hallucination by LLMs \cite{yu2024mechanistic,zhang2023mitigating,shi2024trusting,zhang2023alleviating} work on strategies to eradicate or alleviate the same.  API-based LLMs such as \gpto, \claude are sensitive to the prompts being used and are \textit{variable} in their output generation \cite{zhao2021calibrate}; when we run an instance twice to obtain the two votes necessary to alleviate position bias, the predicted dialog acts for the same human and assistant turns vary slightly across generations as we show in Appendix \ref{app:valid-stats}. A possible solution to reduce hallucinations and variable generation of dialog acts is to train a deterministic classifier model for the same. In Appendix \ref{app:annotation}, we show our qualitative analysis on the correctness of \da (84\%) and \maxim (96\%) annotations by \framework.

\paragraph{Biases in LLMs and mitigation.} We use models such as \gpto, \claude and \qwen that have already been trained (on unknown/unreleased training data) for our experiments; if these models have \textbf{social/demographic biases} \cite{jin-etal-2021-transferability,blodgett-etal-2020-language} (disparate model performance
on different subsets of data which are associated
with different demographic groups) due to their training data, those could get propagated to the models' judgments (for example, ``the better response is $R_1$ because it is of demographic-1 as opposed to demographic-2 that $R_2$ belongs to). Prior works \cite{sun-etal-2019-mitigating,feng-etal-2023-pretraining,gupta-etal-2022-mitigating} propose various frameworks to detect and mitigate specific social biases in language models, \citet{chen2024humans} works on social biases specifically for human and LLM judges. While we don't explicitly work with datasets that necessitate bias mitigation, multi-turn conversations and judging preferences for them are generally fall under this category; furthermore, safety and non-toxicity based preference evaluations (such as the ones needed by the harmless split of anthropic \cite{anthropc-hh-data}) will require bias-free judges. \textbf{Position or order bias} \cite{mtbench,li2024generation} is a prominent issue when using LLMs as judges, wherein the order in which  preference responses are presented to the LLM unfairly influences the LLM-judge's decision. To alleviate this issue, we follow the conservative approach used in prior works \cite{mtbench,wang-etal-2024-large-language-models-fair,hou2024large,qin-etal-2024-large} and present preference responses to the LLM-judge in all possible orders and aggregate all the judgments to make the final decision; we also include in all our prompts a reminder to the model to not be influenced by position bias. \textbf{Verbosity bias.} \cite{mtbench,koo2024benchmarking,park2024offsetbias} is a prominent issue wherein the LLM-judge favors preference responses that are longer/more verbose; we attempt to alleviate this issue by explicitly mentioning in all our prompts a reminder to the model to not be influenced by verbosity.  \textbf{Egocentric bias (self preference).} \cite{li2024generation,koo2024benchmarking} is a bias wherein LLM-judges prefer responses that are generated by the same model as the judge itself. In our work, to the best of our knowledge we do not encounter situations of this nature since all the datasets presumably have responses generated and ranked by multiple language models, and the preference data is \textit{not} GPT versus other models; but ultimately, since we do not know the data used to train \gpto, \claude, \qwen and we do not know the language models used in creating \anthropic, there is a possible that our results are affected by egocentric bias. In general, using LLMs as judges could include more biases such as beauty bias, misinformation oversight bias, authority bias, etc. \cite{li2024generation}.

\paragraph{Trust, Risk.} Trust in a language model is subjective and user-dependent \cite{lipton2018mythos}; some users trust models that provide a high accuracy, some users require the model's reasoning process to be clear, etc. Further, risks such as hallucination, bias, etc. are also pivotal in the debate about trust. As we discuss above, we work on reducing hallucinations and bias in our judges and also propose judges and juries that improve the judgment accuracy; however, we still urge any researcher/user who utilizes our methods to qualitatively analyze our methods' predictions to ensure quality and trust. 


\section*{Reproducibility}

\paragraph{Data and models.} All the datasets that we use in our work are released publicly for usage and have been duly attributed to their original authors. We do not train or release any models; we release all prompts in Appendix \ref{app:io-wexpl-prompts}, \ref{app:da-prompt}, \ref{app:maxim-prompt}. We use publicly available models \gptofull, \claudefull, \qwenfull, \skyllamafull, \qrmfull, \informfull (evaluation parameters in Appendix \ref{app:reproducibility}). We perform only evaluation and no training; all our evaluation dataset details are in Appendix \ref{app:data-cleaning}. All code and data are uploaded at \href{https://github.com/INK-USC/Amulet-DA-and-Maxim}{\texttt{github.com/INK-USC/Amulet-DA-and-Maxim}}. 

\paragraph{Usage and License.} All the datasets we use are available open-sourced on hugging-face: \href{https://huggingface.co/datasets/Anthropic/hh-rlhf}{\anthropic}, \href{https://huggingface.co/datasets/microsoft/WildFeedback}{\wildf}, \href{https://huggingface.co/datasets/allenai/tulu-2.5-preference-data}{\nectar}. All datasets are licensed to be freely used and distributed (\nectar alone is licensed to not compete with OpenAI, which we comply with). All these datasets are anonymized; we do not separately take steps to identify and remove offensive data.  
We use \gpto and \claude as paid API services with corresponding API keys. We use the following models from hugging-face: \href{https://huggingface.co/Skywork/Skywork-Reward-Llama-3.1-8B-v0.2}{\skyllamafull}, \href{https://huggingface.co/nicolinho/QRM-Gemma-2-27B}{\qrmfull},    \href{https://huggingface.co/infly/INF-ORM-Llama3.1-70B}{\informfull}, \href{https://huggingface.co/R-I-S-E/RISE-Judge-Qwen2.5-32B}{\qwenfull}. All these models are open-sourced; to access \qwen, we had to provide our contact information and accept the terms and conditions. All data and models were used consistent with their intended use.



\section*{Acknowledgements}
This research is supported in part by the Office of the Director of National Intelligence (ODNI), Intelligence Advanced Research Projects Activity (IARPA), via the HIATUS Program contract \#2022-22072200006. The views and conclusions contained herein are those of the authors and should not be interpreted as necessarily representing the official policies, either expressed or implied, of ODNI, IARPA, or the U.S. Government. The U.S. Government is authorized to reproduce and distribute reprints for governmental purposes notwithstanding any copyright annotation therein. Brihi Joshi was supported by the Apple Scholars in AI/ML PhD fellowship. Finally, we thank all lab members at USC INK Research Lab for their constructive feedback on this work.

\bibliography{custom}

\appendix

\section{Related Work}
\paragraph{LLM judges.} Prior work on text quality and dialog metrics include model-free metrics such BLEU, ROUGE, engaging-ness (based on length of conversation) etc.  and metrics with trained models such as grammar, fluency, coherence, error recovery, informativeness, likability, inquisitiveness, etc. \cite{papineni2002bleu,pang-etal-2020-towards,yeh-etal-2021-comprehensive,fu2024gptscore,mehri2022report,zhang2022fined,ni2023recent}. Today, with the increasing proficiency of language models, it has become harder to measure the quality of the text generated by them with standardized metrics; human evaluation of these responses are too expensive and time-consuming \cite{ouyang2022training,mtbench}. This has led to the adoption of large language models as judges \cite{mtbench}; these LLM-judges mimic human judgments and preferences, and are the most commonly used evaluators today. 

LLM-judges can provide feedback in various forms \cite{mtbench,wildbench}: (1) preference/rankings for multiple texts, which can serve as either just evaluation or as RL-AIF data as well \cite{cui2023ultrafeedback}, (2) likert/numerical scores that indicate the quality of the text (the judge's prompt can specify exact score levels or reasoning, or can just request for a score between say 1-10), (3) natural language feedback that can be used to improve language models \cite{akyurek-etal-2023-rl4f}.

Current work in LLM-judges also include juries of judges \cite{wildbench,verga2024replacing,jung2024trust}, training smaller language models to be judges \cite{mtbench,zhujudgelm,kim2023prometheus}. 

\paragraph{Single-turn and multi-turn.} Contemporary works on benchmarking generative language models and reward models largely focus on training and analysis with single- or two-turn datasets, for example, UltraFeedback \cite{cui2023ultrafeedback}, SHP \cite{pmlr-v162-ethayarajh22a}, ChatArena \cite{chatbotarena}, MT-Bench \cite{mtbench}, Alpaca-Eval \cite{alpaca_eval}, RewardBench \cite{rewardbench} etc. A recent benchmark WildBench \cite{wildbench} takes a step in this direction and focuses on multi-turn data. But WildBench only works with $\leq 5$ human turns of data and has only 1024 evaluation instances; further WildBench heavily focuses on singular task based examples (such as math, editing, data analysis). Other recent preference datasets include PRISM \cite{prism} and Argilla-Capybara \cite{capybara}: unfortunately, PRISM includes many instances where the chosen and rejected responses are highly similar, and Argilla-Capybara is completely synthetic and heavily based on individual tasks; hence, we are unable to use these in our multi-turn preference benchmark. More recently, multi-turn datasets such as WildChat \cite{zhao2024wildchat1mchatgptinteraction}, LMSYS-Chat \cite{zhenglmsys} etc. which contain real-life conversations between humans and various LM assistants have been released - unfortunately, these datasets do not have preference information and can only be used for supervised fine-tuning. 


\paragraph{Dialog Metrics and Grice's maxims.} Contemporary works specify varied metrics that a language model has to satisfy to be a good conversational agent; these include metrics pertaining to conversational AI such as flexibility to the human user, user satisfaction, error recovery, as well as metrics pertaining to language usage such as grammar, fluency, factuality \cite{ni2023recent,zhang2022fined,mehri2022report,fu2024gptscore,yeh-etal-2021-comprehensive}. However, not all these metrics have clear definitions, and there exists no \textit{complete} list of metrics that are sufficient and necessary for a language model to be considered universally good at making conversation. Hence, in our analysis of human-assistant conversations, to measure the properties that the AI assistant (dis)satisfies, we go back to the fundamental conversational principles of \textit{Gricean Maxims} \cite{grice1975logic}. Grice's maxims have been used in NLP literature in varying degrees: \citet{krause2024gricean} designates the most important maxims for different NLP tasks, \citet{kasirzadeh2023conversation} discusses Grice's maxims in terms of various domains. However, at the same time, there have also been discussions in linguistics regarding the vagueness of the maxim definitions \cite{frederking1996grice} and the insufficiency of Grice's maxims with respect to (1) non-cooperative communication such as irony and sarcasm, (2) adaptability to and extension across cultures  \cite{hossain2021application}.

\paragraph{Reward models.} Since its release, RewardBench \cite{rewardbench} has served as a solid benchmark for reward models, and has standardized the evaluation of reward models. Most of these reward models are classifiers \cite{inf-orm,qrm,skywork,armorm,urm}, producing an unbounded numerical score; the higher the score, the better the quality of the text being analyzed. RewardBench largely focuses on single-turn conversations. 

\section{Example complex multi-turn conversation} \label{app:example-complex-mt}
As a first step towards understanding the complexity of multi-turn conversations, we manually analyzed several conversations from datasets including MT-Bench \cite{mtbench}, WildChat \cite{zhao2024wildchat1mchatgptinteraction}, LMSYS-Chat \cite{zhenglmsys}. We observed that the user intents often change from one turn to the next. We preliminarily observed the following intents (sample in Table \ref{tab:complex-sample-appendix}): social (greeting/thanking/goodbye), role playing instructions, asking the assistant to perform a task, seeking information from the assistant, stating a fact, giving additional/relevant information about a situation, expressing an emotion, technical/error statements (which implicitly request the assistant to explain or debug the error). We used this as our motivation and sought to formalize the changing intents in a conversation; we adapt and present \textit{dialog acts} from \citet{bunt2019guidelines} that explain the communicative structures and intents in every turn of a conversation.

\begin{table*}[ht]
    \centering
    \small
    \begin{tabular}{p{0.6\textwidth}|p{0.38\textwidth}}
    \hline
    \textbf{Human turn} & \textbf{Observed Intents} \\ \hline \hline
    \hl{Example-1} & - \\ \hline
    Have you ever heard of the National Grid in the UK? & seeking information  \\ \hline
    Can you tell me where the National Grid Control Center is? & seeking information, asking the assistant to perform a task \\ \hline
    Do you have the exact address. I need to deliver an item there & seeking information, asking the assistant to perform a task, giving additional/relevant information about a situation \\ \hline
    Maye you have its post code? & seeking information, asking the assistant to perform a task \\ \hline
    Oh that's strange I thought it began with RG41? & expressing an emotion, stating a fact, giving additional/relevant information about a situation \\ \hline \hline
    \hl{Example-2} & - \\ \hline
    I'm feeling sad. Can you tell me a joke to cheer me up? & expressing an emotion, asking the assistant to perform a task \\ \hline
    My grandmother used to tell me jokes about Irish people, which always cheered me up. Pretend to be my grandmother and cheer me up & role-playing instructions, asking the assistant to perform a task \\ \hline
    My grandmother used to make jokes about how Irish people have no potatoes. How about a potato joke, something like: Grandma: what's the difference between an Irishman and a bench? & giving additional/relevant information about a situation, asking the assistant to perform a task \\ \hline
    \end{tabular}
    \caption{Consecutive human turns from a conversation in WildChat \cite{zhao2024wildchat1mchatgptinteraction}. We observe multiple intents in almost all the turns, and also observe that the intent of the human keeps changing over the turns.}
    \label{tab:complex-sample-appendix}
\end{table*}

\section{\da and \maxim} \label{app:da-and-maxim}
Tables \ref{tab:dialog-acts} and \ref{tab:maxims} provide a full list of all dialog acts and maxims we use in this work. We obtain the dialog act taxonomy from \citet{bunt2019guidelines} and the maxim definitions from \citet{grice1975logic,miehling2024language}.
\begin{table*}[ht]
    \centering
    \small
    \begin{tabular}{|p{0.3\textwidth}|p{0.3\textwidth}|p{0.38\textwidth}|}
    \toprule
    \textbf{Communicative Dimension} & \textbf{Definition of Communicative Dimension} & \textbf{Communicative Function} \\ \midrule 
    Task & Underlying task/activity & Questions, Answers, Commissives, Directives \\  \midrule
    Auto-feedback & Speaker’s processing of previous utterances & Positive/negative feedback \\ \midrule
    Allo-feedback & Addressee’s processing of previous utterances & Positive/negative feedback \\ \midrule
    Turn Management & Allocation of turn [implicit, mostly] & Keep/grab/give turn \\  \midrule
    Time Management & Time needed to speak & Stalling, Pausing \\ \midrule
    Contact Management & Check contact & Check if contact is there \\ \midrule
    Own Communication Management & Editing speaker’s own turn & Self-correction \\ \midrule
    Partner Communication Management & Editing what the addressee said & Correct misspeaking / Completion of addressee \\ \midrule
    Discourse/Interaction Structuring & Structure of dialogue & New topic, open/close dialogue, announce change in structure \\ \midrule
    Social Obligations Management & Social obligation & Greeting, Thanks, Apology and corresponding responses \\ \bottomrule
    \end{tabular}
    \caption{\textbf{Dialog Acts:} Communicative dimensions and functions from \citet{bunt2019guidelines}}
    \label{tab:dialog-acts}
\end{table*}

\begin{table*}[ht]
    \centering
    \small
    \begin{tabular}{|p{0.25\textwidth}|p{0.7\textwidth}|}
    \toprule
    \textbf{Maxim} & \textbf{Definition} \\ \midrule
    Quantity &  Related to the informativity, that is, the amount of information contained in a given response. \\ \midrule
    Quality &  Pertains to the truthfulness and honesty of the response \\ \midrule
    Relation & The relevance of the response to the conversation \\ \midrule
    Manner & Logical flow and clarity of the discourse \\ \midrule
    Benevolence & Moral responsibility of the response, particularly concerning the generation of, and engagement with, harmful content \\ \midrule
    Transparency & concerning recognition of one’s knowledge boundaries, operational constraints, and intents \\ \midrule \midrule
    \cellcolor{lblue} Quantity-1 & The response should provide a sufficient amount of information. \\ \midrule
    \cellcolor{lblue} Quantity-2 & The response should not contain unnecessary details. \\ \midrule
    \cellcolor{lblue} Quality & The response should be factual and supported by adequate evidence whenever possible. \\ \midrule
    \cellcolor{lblue} Relevance-1 & The response should directly and relevantly address the recipient's statements in a helpful manner. \\ \midrule
    \cellcolor{lblue} Relevance-2 &  The response should be relevant to the current topic and not unnaturally shift the conversation to unrelated subjects. \\ \midrule
    \cellcolor{lblue} Manner-1 & The response should be clear, unambiguous, and presented in a well-organized fashion. \\ \midrule
    \cellcolor{lblue} Manner-2 & The response should be accessible and use appropriate language tailored to the recipient's level of understanding. \\ \midrule
    \cellcolor{lblue} Benevolence-1 & The response should not exhibit insensitivity, rudeness, or harm. \\ \midrule
    \cellcolor{lblue} Benevolence-2 & The response should not reflect an engagement or endorsement with requests that are harmful or unethical. \\ \midrule
    \cellcolor{lblue} Transparency-1 & The response should recognize the speaker's knowledge boundaries, making clear any limitations in expertise, evidence, experience, or context. \\ \midrule
    \cellcolor{lblue} Transparency-2 & The response should recognize the speaker's operational capabilities, highlighting the nature of actions that can or cannot be performed. \\ \midrule
    \cellcolor{lblue}  Transparency-3 & The response should be forthright about the speaker's willingness to engage with specific subjects or heed relevant advice.     \\ \bottomrule
    \end{tabular}
    \caption{\textbf{Maxims:} Definitions quoted from \citet{miehling2024language}. The highlighted cells are the sub-maxims we use in \framework-\maxim.}
    \label{tab:maxims}
\end{table*}

\section{Data cleaning} \label{app:data-cleaning}
In this section, we expand upon our data cleaning strategies. 
\begin{itemize}
    \item Step-1: Only retain conversations with $\geq 4$ human turns.
    \item Step-2: Remove all instances where the chosen and rejected responses are \textit{exactly} the same (done via string matching).
    \item Step-4: Remove all instances where the structure of the instance is ill-formed, for example, two human turns in a row.
    \item Step for \nectar: We only consider instances from the first $100000$ instances in \nectar in order to control the size of the evaluation dataset.
    \item Step for \nectar: Since \nectar has been created from \anthrtrain, we check for and remove the following instances: (1) instances also present in \anthrtrain, (2) instances where the chosen and rejected responses are opposite to that of \anthrtrain
    \item Step for \wildf: We only retain instances in \wildf where every human or assistant turn is less than $300$ words in length.
\end{itemize}

The final evaluation set sizes are as follows: \anthrtest - 460, \anthrtrain - 8210, \nectar - 6531 and \wildf - 4684.\\

The evaluation set sizes for conversations with $\geq 7$ turns are as follows:  \anthrtrain - 548, \nectar - 335 and \wildf - 1867.\\

\section{\io and \wexpl prompts} \label{app:io-wexpl-prompts}
We present the baseline judge \io and \wexpl prompts in Tables \ref{tab:io-prompt} and \ref{tab:wexpl-prompt} respectively.
\begin{table*}[h!]
    \centering
    \small
    \begin{tabular}{p{0.98\textwidth}}
    \toprule
    \textbf{Instruction:} You will be given a dialog conversation between a human user and an LLM assistant. The dialog is split into turns - a turn is defined as an utterance by either the human user or the assistant. Note that the roles of ``speaker'' (S) and ``addressee'' (A) will alternate at every turn. The last turn alone will have two responses, sampled from different LLM assistants. Your task is to analyze the two responses and say which one of them is better - you should take all the previous turns of the dialog into consideration. Your evaluation should consider factors such as the helpfulness, relevance, accuracy, depth, creativity, and level of detail of their responses. Avoid any position biases and ensure that the order in which the responses were presented does not influence your decision. Do not allow the length of the responses to influence your evaluation. Be as objective as possible. An example is given below with the required JSON output format. Say ``1'' if you think Assistant-1's response is better and ``2'' if you think Assistant-2's response is better. \\\\
    \textbf{Example Dialog - }\\
    \textbf{Human:} Human's turn \\
    \textbf{Assistant:} Assistant's turn \\
    \textbf{Human:} Human's turn \\
    \textbf{Assistant:} Assistant's turn \\
    \textbf{Human:} Human's turn \\
    \textbf{Assistant-1:} \hl{Assistant's turn by Assistant-1} \\
    \textbf{Assistant-2:} \hl{Assistant's turn by Assistant-2} \\\\
    \textbf{Example Output -} \\
    \{\\
    	\textbf{``Answer'':} \hl{``fill either 1 or 2''},\\
    \}\\
    \bottomrule
    \end{tabular}
    \caption{Prompt for \io}
    \label{tab:io-prompt}
      
\end{table*}

\begin{table*}[h!]
    \centering
    \small
    \begin{tabular}{p{0.98\textwidth}}
    \toprule
    \textbf{Instruction:} You will be given a dialog conversation between a human user and an LLM assistant. The dialog is split into turns - a turn is defined as an utterance by either the human user or the assistant. Note that the roles of ``speaker'' (S) and ``addressee'' (A) will alternate at every turn. The last turn alone will have two responses, sampled from different LLM assistants. Your task is to analyze the two responses and say which one of them is better - you should take all the previous turns of the dialog into consideration. Your evaluation should consider factors such as the helpfulness, relevance, accuracy, depth, creativity, and level of detail of their responses. Avoid any position biases and ensure that the order in which the responses were presented does not influence your decision. Do not allow the length of the responses to influence your evaluation. Be as objective as possible. An example is given below with the required JSON output format. Say ``1'' if you think Assistant-1's response is better and ``2'' if you think Assistant-2's response is better. \\\\
    \textbf{Example Dialog - }\\
    \textbf{Human:} Human's turn \\
    \textbf{Assistant:} Assistant's turn \\
    \textbf{Human:} Human's turn \\
    \textbf{Assistant:} Assistant's turn \\
    \textbf{Human:} Human's turn \\
    \textbf{Assistant-1:} \hl{Assistant's turn by Assistant-1} \\
    \textbf{Assistant-2:} \hl{Assistant's turn by Assistant-2} \\\\
    \textbf{Example Output -} \\
    \{\\
    	\textbf{``Answer'':} \hl{``fill either 1 or 2''},\\
    	\textbf{``Explanation'':} \hl{``fill your explanation here''}\\
    \}\\
    \bottomrule
    \end{tabular}
    \caption{Prompt for \wexpl}
    \label{tab:wexpl-prompt}
      
\end{table*}

\section{The \framework-\da prompt} \label{app:da-prompt}

This section contains the prompt we use for \framework-\da with \texttt{\textbf{this font}}; we provide this prompt outside of a table since it is too big to be accommodated in a table.

\noindent \texttt{\textbf{Instruction:} You will be given a dialog conversation between a human user and an LLM assistant. The dialog is split into turns - a turn is defined as an utterance by either the human user or the assistant. Note that the roles of ``speaker'' (S) and ``addressee'' (A) will alternate at every turn. The last turn alone will have two responses, sampled from different LLM assistants. \textbf{Your task is to label each turn of dialogue in terms of dialog-acts - dialog acts are defined in terms of communicative dimensions ``Dim'' and corresponding communicative functions ``Func''} [detailed below along with their meanings]. For each turn of dialogue, you must mark *all* the dimensions and functions that are present. You should take the previous turns of the dialog into consideration when labeling the dialog acts. \textbf{Finally, use these dialog acts to determine which response is better} - say ``1'' if you think Assistant-1's response is better and ``2'' if you think Assistant-2's response is better. Also provide an explanation for your choice. \textbf{Avoid any position biases and ensure that the order in which the responses were presented does not influence your decision. Do not allow the length of the responses to influence your evaluation. Be as objective as possible.} Examples are given below with the desired output format.\\\\}
\texttt{List of dimensions and functions - \\  
\textbf{(1) Dim: Task}, Meaning: Underlying task/activity, \textbf{Func:} Propositional Question, Set Question, Choice Question, Answer, Confirm, Disconfirm, Inform, Agreement, Disagreement, Correction, Promise, Offer, Accept Request, Decline Request, Accept Suggest, Decline Suggest, Request, Instruct, Suggest \\
\textbf{(2) Dim: Auto-Feedback}, Meaning: Speaker's processing of previous utterances, \textbf{Func:} Auto-Positive, Auto-Negative \\
\textbf{(3) Dim: Allo-Feedback}, Meaning: Speaker expressing opinions about addressee's processing of previous utterances, \textbf{Func:} Allo-Positive, Allo-Negative, Feedback Elicitation \\
\textbf{(4) Dim: Time Management}, Meaning: Concerning the allocation of time to the speaker, \textbf{Func:} Stalling, Pausing \\
\textbf{(6) Dim: Own Communication Management}, Meaning: Editing speaker's own speech within the current turn, \textbf{Func:} Self-Correction, Self-Error, Retraction \\
\textbf{(7) Dim: Partner Communication Management}, Meaning: Editing what the addressee said, \textbf{Func:} Completion, Correct Misspeaking,  \\
\textbf{(8) Dim: Discourse/Interaction Structuring}, Meaning: Explicitly structuring the interaction, \textbf{Func:} Interaction Structuring, Opening, Closing \\
\textbf{(9) Dim: Social Obligations Management}, Meaning: Social obligation, \textbf{Func:} Initial Greeting, Return Greeting, Intitial Self-Introduction, Return Self-Introduction, Apology, Accept Apology, Thanking, Accept Thanking, Initial Goodbye, Return Goodbye \\
You should use only the above dimensions and functions, do not make up new ones. Below you will find detailed definitions of the functions. \textbf{To reiterate, S refers to the speaker and A refers to the addressee. Both the human and the LM assistant take on the roles of S and A in alternating turns.}}\\\\
\texttt{
\textbf{Information-seeking functions - } \\
$*$ Propositional Question: Function performed by S, in order to know whether the proposition, which forms the semantic content, is true. S assumes that A knows whether the proposition is true or not, and puts pressure on A to provide this information. \\
$*$ Set Question: Function performed by S, in order to know which elements of a given set have a certain property specified by the semantic content; S puts pressure on A to provide this information, which S assumes that A possesses. S believes that at least one element of the set has that property. \\
$*$ Choice Question: Function performed by S, in order to know which one from a list of alternative propositions, specified by the semantic content, is true; S believes that exactly one element of that list is true; S assumes that A knows which of the alternative propositions is true, and S puts pressure on A to provide this information. \\\\
\textbf{Information-providing functions - }\\
$*$ Answer: Function performed by S, in order to make certain information available to A which S believes A wants to know; S assumes that this information is correct. \\
$*$ Confirm: Function performed by S, in order to inform A that certain information that A wants to know, and concerning which A holds an uncertain belief, is indeed correct. \\
$*$ Disconfirm: Function performed by S, in order to let A know that certain information that A wants to know, and concerning which A holds an uncertain belief, is incorrect. \\
$*$ Inform: Function performed by S, in order to make the information contained in the semantic content known to A; S assumes that the information is correct. \\
$*$ Agreement: Function performed by S, in order to inform A that S assumes a given proposition to be true, which S believes that A also assumes to be true. \\
$*$ Disagreement: Function performed by S, in order to inform A that S assumes a given proposition to be false, which S believes that A assumes to be true. \\
$*$ Correction: Function performed by S, in order to inform A that certain information which S has reason to believe that A assumes to be correct, is in fact incorrect and that instead the information that S provides is correct. \\\\
\textbf{Commissive functions - } \\
$*$ Promise: Function by which S commits to perform the action specified in the semantic content, in the manner or with the frequency or depending on the conditions that S makes explicit. S believes that this action would be in A's interest. \\
$*$ Offer: Function by which S indicates willingness and ability to perform the action specified by the semantic content, conditional on the consent of A that S do so. \\
$*$ Accept Request: Function by which S commits to perform an action that S has been requested to perform, possibly depending on certain conditions that S makes explicit. \\
$*$ Decline Request: Function by which S refuses to perform an action that S has been requested to perform, possibly depending on certain conditions that S makes explicit. \\
$*$ Accept Suggest: Function by which S commits to perform an action that was suggested, possibly with certain restrictions or conditions concerning manner or frequency of performance. \\
$*$ Decline Suggest: Function by which S indicates that S will not perform an action that was suggested, possibly depending on certain conditions that S makes explicit. \\\\
\textbf{Directive functions - }\\
$*$ Request: Function performed by S, in order to create a commitment for A to perform a certain action in the manner or with the frequency described by the semantic content, conditional on A's consent to perform the action. S assumes that A is able to perform this action. \\
$*$ Instruct: Function performed by S, in order to create a commitment for A to carry out a named action in the manner or with the frequency specified by the semantic content; S assumes that A is able and willing to carry out the action. \\
$*$ Suggest: Function performed by S, in order to make A consider the performance of a certain action specified by the semantic content. S believes that this action is in A's interest, and assumes that A is able to perform the action.}\\\\
\texttt{
\textbf{Feedback functions -} \\ 
$*$ Auto-Positive: Function performed by S, in order to inform A that S believes that S's processing of the previous utterance(s) was successful. \\
$*$ Allo-Positive: Function performed by S, in order to inform A that S believes that A's processing of the previous utterance(s) was successful. \\
$*$ Auto-Negative: Function performed by S, in order to inform A that S's processing of the previous utterance(s) encountered a problem. \\
$*$ Allo-Negative: Function performed by S, in order to inform A that S believes that A's processing of the previous utterance(s) encountered a problem. \\
$*$ Feedback Elicitation: Function performed by S, in order to know whether A's processing of the previous utterance(s) was successful.\\\\
\textbf{Time management functions - }\\
$*$ Stalling: Function performed by S, in order to have a little extra time to construct S's contribution. \\
$*$ Pausing: Function performed by S, in order to suspend the dialogue for a short while. \\\\
\textbf{Own and Partner Communication Management Functions -} \\
$*$ Completion: Function performed by S in order to assist A in the completion of an utterance. \\
$*$ Correct Misspeaking: Function performed by S, in order to correct (part of) an utterance by A assuming that A made a speaking error. \\
$*$ Self-Error: Function performed by S, in order to signal to the A that S has made a mistake in speaking. \\
$*$ Retraction: Function performed by S, in order to withdraw something that S just said within the same turn. \\
$*$ Self-Correction: Function performed by S, in order to correct an error that S just made, or to improve on an infelicitous formulation that S just used, within the same turn.\\\\
\textbf{Discourse structuring functions - }\\
$*$ Interaction Structuring: Function performed by S, in order to explicitly indicate to A the function or topic of S's next contribution(s). \\
$*$ Opening: Function performed by S, in order to inform A that S is ready and willing to engage in a dialogue with A.\\
$*$ Closing: Function performed by S, in order to inform A that S is about to end the conversation.\\\\
\textbf{Social obligations management functions - } \\
 $*$ Initial Greeting: Function performed by S, in order to inform A that S is present and aware of A's presence; S puts pressure on A to acknowledge this. \\
$*$ Return Greeting: Function performed by S, in order to acknowledge that S is aware of A's presence, and of A having signalled A's presence to S.
$*$ Initial Self-Introduction: Function performed by S, in order to make A know about S; S puts pressure on A to acknowledge this. \\
$*$ Return Self-Introduction: Function performed by S, in order to make A know about S in response to a self-introduction by A. \\
$*$ Apology: Function performed by S, that signals S wants A to knows that S regrets something; S puts pressure on A to acknowledge this. \\
$*$ Accept Apology: Function performed by S, in order to mitigate the feelings of regret that A has expressed. \\
$*$ Thanking: Function performed by S, in order to inform A that S is grateful for some action performed by A; S puts pressure on A to acknowledge this. \\
$*$ Accept Thanking: Function performed by S, in order to mitigate the feelings of gratitude which A has expressed. \\
$*$ Initial Goodbye: Function performed by S, in order to inform A that S intends the current utterance to be S's final contribution to the dialogue; S puts pressure on A to acknowledge this.\\
$*$ Return Goodbye: Function performed by S, in order to acknowledge awareness that A has signalled A's final contribution to the dialogue and S signals in return S's agreement to end the dialogue; S has been pressured to respond to an initial goodbye by A.}\\\\
\texttt{Important additional note: If a turn contains the answer to a question from a prior turn, make sure to mark the communicative function ``Answer'' in addition to any other function present in the turn (such as ``Inform'', ``Suggest'', ``Instruct'', etc.). That is, even if the answer is phrased as an instruction, or a suggestion, etc., if the turn includes the answer to a question from a prior turn, specify ``Answer'' as one of the functions in the turn.}\\\\
\texttt{
\textbf{Example Dialog 1 - }\\
\textbf{Human:} How do I teach kids to meditate? \\
\textbf{Assistant:} Great question! That's a really useful skill to cultivate, it can bring peace, calm, and happiness. I'm glad you want to teach your kids about it. \\
\textbf{Human:} All right, so how do we start? \\
\textbf{Assistant:} Well, we can get started with just being silent. You can tell the kids it's okay if they just sit there quietly for a few minutes without thinking of anything.
\textbf{Human:} any other ideas? they are fidgeting \\
\textbf{Assistant-1:} Kids are great at fidgeting, so let's make sure we give them permission to fidget. \\
\textbf{Assistant-2:} Yeah, definitely. What works for you personally? What works for other people?\\\\
\textbf{Example Output 1 - }\\
\textbf{Human:} How do I teach kids to meditate? <SEP> \{``Dim'': ``Task'', ``Func'': ``Set Question''\} \\
\textbf{Assistant:} Great question! That's a really useful skill to cultivate, it can bring peace, calm, and happiness. I'm glad you want to teach your kids about it. <SEP> \{``Dim'': ``Allo-Feedback'', ``Func'': ``Allo-Positive''\} \{``Dim'': ``Task'', ``Func'': ``Inform''\}   \\
\textbf{Human:} All right, so how do we start? <SEP> \{``Dim'': ``Allo-Feedback'', ``Func'': ``Allo-Positive''\} \{``Dim'': ``Task'', ``Func'': ``Request \& Set Question''\} \\
\textbf{Assistant:} Well, we can get started with just being silent. You can tell the kids it's okay if they just sit there quietly for a few minutes without thinking of anything. <SEP> \{``Dim'': ``Time Management'', ``Func'': ``Stalling''\} \{``Dim'': ``Task'', ``Func'': ``Answer''\} \\
\textbf{Human:} any other ideas? they are fidgeting <SEP> \{``Dim'': ``Allo-Feedback'', ``Func'': ``Allo-Negative''\} \{``Dim'': ``Task'', ``Func'': ``Set Question \& Inform''\} \\
\textbf{Assistant-1:} Kids are great at fidgeting, so let's make sure we give them permission to fidget. <SEP> \{``Dim'': ``Task'', ``Func'': ``Inform \& Instruct''\} \\
\textbf{Assistant-2:} Yeah, definitely. What works for you personally? What works for other people? <SEP> ````Dim'': ``Auto-Feedback'', ``Func'': ``Auto-Positive''\} \{``Dim'': ``Task'', ``Func'': ``Set Question''\}\\\\
\{\\
	``\textbf{Answer}'': ``fill either 1 or 2'',\\
	``\textbf{Explanation}'': ``fill your explanation here''\\
\}\\\\
\textbf{Example Dialog 2 - } \\
\textbf{Human:} what are some good novels for a 9 year old? \\
\textbf{Assistant:} That depends on the 9 year old, but if they like thrillers or mysteries then maybe something like ``The Body in the Bog'' or ``The Boy Who Followed Ripley.'' For more modern, female-written fiction I might recommend ``Jane Eyre'' or ``Frankenstein.'' \\
\textbf{Human:} That sounds cool \\
\textbf{Assistant-1:} I'm a programmer, so I'm really into good programming texts, like ``The Practice of Programming'' or ``JavaScript: The Good Parts.'' But that might be a bit too much for a 9 year old, so \\
\textbf{Assistant-2:} Do you want some quotes?\\\\
\textbf{Example Output 2 - } \\
\textbf{Human:} what are some good novels for a 9 year old? <SEP> {``Dim'': ``Task'', ``Func'': ``Set Question''} \\
\textbf{Assistant:} That depends on the 9 year old, but if they like thrillers or mysteries then maybe something like ``The Body in the Bog'' or ``The Boy Who Followed Ripley.'' For more modern, female-written fiction I might recommend ``Jane Eyre'' or ``Frankenstein.'' <SEP> \{``Dim'': ``Answer \& Suggest''\} \\
\textbf{Human:} That sounds cool <SEP> \{``Dim'': ``Allo-Feedback'', ``Func'': ``Allo-Positive''\} \\
\textbf{Assistant-1:} I'm a programmer, so I'm really into good programming texts, like ``The Practice of Programming'' or ``JavaScript: The Good Parts.'' But that might be a bit too much for a 9 year old, so <SEP> \{``Dim'': ``Task'', ``Func'': ``Inform''\} \{``Dim'': ``Own Communication Management'', ``Func'': ``Retraction''\} \\
\textbf{Assistant-2:} Do you want some quotes? <SEP> \{``Dim'': ``Task'', ``Func'': ``Offer \& Propositional Question''\}\\\\
\{\\
	``\textbf{Answer}'': ``fill either 1 or 2'',\\
	``\textbf{Explanation}'': ``fill your explanation here''\\
\} \\\\
\textbf{Dialog -}} 

\subsection{The \framework-\da prompt for \claude}
With \claude, we encountered a unique problem - \claude stopped its generation after the dialog act for the first human turn because it assumed that its turn was \textit{over}. Hence, we use the following modified output format for the few-shot examples in \framework-\da's prompt for \claude:\\\\
\texttt{\textbf{Example Output 1 -}\\
\{\\
````\textbf{Human:} How do I teach kids to meditate?'''': ````\{``Dim'': ``Task'', ``Func'': ``Set Question''\}'''', \\
````\textbf{Assistant:} Great question! That's a really useful skill to cultivate, it can bring peace, calm, and happiness. I'm glad you want to teach your kids about it.'''': ````\{``Dim'': ``Allo-Feedback'', ``Func'': ``Allo-Positive'\} \{``Dim'': ``Task'', ``Func'': ``Inform''\}'''', \\
````\textbf{Human:} All right, so how do we start?'''': ````\{``Dim'': ``Allo-Feedback'', ``Func'': ``Allo-Positive'\} \{``Dim'': ``Task'', ``Func'': ``Request \& Set Question''\}'''', \\
````\textbf{Assistant:} Well, we can get started with just being silent. You can tell the kids it's okay if they just sit there quietly for a few minutes without thinking of anything.'''': ````\{``Dim'': ``Time Management'', ``Func'': ``Stalling''\} \{``Dim'': ``Task'', ``Func'': ``Answer''\}'''', \\
````\textbf{Human:} any other ideas? they are fidgeting'''': ````\{``Dim'': ``Allo-Feedback'', ``Func'': ``Allo-Negative''\} \{``Dim'': ``Task'', ``Func'': ``Set Question \& Inform''\}'''', \\
````\textbf{Assistant-1:} Kids are great at fidgeting, so let's make sure we give them permission to fidget.'''': ````\{``Dim'': ``Task'', ``Func'': ``Inform \& Instruct''\}'''', \\
````\textbf{Assistant-2:} Yeah, definitely. What works for you personally? What works for other people?'''': ````\{``Dim'': ``Auto-Feedback'', ``Func'': ``Auto-Positive''\} \{``Dim'': ``Task'', ``Func'': ``Set Question''\}'''',\\
	`\textbf{Answer}': `fill either 1 or 2',\\
	`\textbf{Explanation}': `fill your explanation here'\\
\}
}

\section{The \framework-\maxim prompt} \label{app:maxim-prompt}

This section contains the prompt we use for \framework-\maxim with \texttt{\textbf{this font}}; we provide this prompt outside of a table since it is too big to be accommodated in a table.

\texttt{
\textbf{Instruction:} You will be given a dialog conversation between a human user and an LLM assistant. The dialog is split into turns - a turn is defined as an utterance by either the human user or the assistant. Note that the roles of ``speaker'' and ``addressee'' will alternate at every turn. The last turn alone will have two responses, sampled from different LLM assistants.  Your task is to analyze the two responses and say which one of them is better using the below maxim based evaluation protocol. 
For each of the twelve maxims described below, indicate whether response 1 (from Assistant-1) or response 2 (from Assistant-2) is better in satisfying the maxim by specifying ``1'' or ``2'' respectively. If both responses satisfy the maxim equally, say ``both'' and if neither response satisfies the maxim, say ``neither''. Finally provide a free-text explanation for your decisions and provide your answer of whether response 1 or response 2 is better. You should take the previous turns of the dialog into consideration when labeling the satisfaction of maxims and when providing your final answer. Avoid any position biases and ensure that the order in which the responses were presented does not influence your decision. Do not allow the length of the responses to influence your evaluation. Be as objective as possible. An example is given below with the required JSON output format. Say ``1'' if you think Assistant-1's response is better and ``2'' if you think Assistant-2's response is better. You have to pick ``1'' or ``2'', you cannot say ``both'' for the final answer.\\\\
List of Grice's maxims -  \\
\textbf{(1) Quantity-1:} The response should provide a sufficient amount of information. \\
\textbf{(2) Quantity-2:} The response should not contain unnecessary details. \\
\textbf{(3) Quality:} The response should be factual and supported by adequate evidence whenever possible. \\
\textbf{(4) Relevance-1:} The response should directly and relevantly address the recipient's statements in a helpful manner. \\
\textbf{(5) Relevance-2:}  The response should be relevant to the current topic and not unnaturally shift the conversation to unrelated subjects. \\
\textbf{(6) Manner-1:} The response should be clear, unambiguous, and presented in a well-organized fashion. \\
\textbf{(7) Manner-2:} The response should be accessible and use appropriate language tailored to the recipient's level of understanding. \\
\textbf{(8) Benevolence-1:} The response should not exhibit insensitivity, rudeness, or harm. \\
\textbf{(9) Benevolence-2:} The response should not reflect an engagement or endorsement with requests that are harmful or unethical. \\
\textbf{(10) Transparency-1:} The response should recognize the speaker's knowledge boundaries, making clear any limitations in expertise, evidence, experience, or context. \\
\textbf{(11) Transparency-2:} The response should recognize the speaker's operational capabilities, highlighting the nature of actions that can or cannot be performed. \\
\textbf{(12) Transparency-3:} The response should be forthright about the speaker's willingness to engage with specific subjects or heed relevant advice. \\\\
You should use only the above maxims, do not make up new ones.\\\\
\textbf{Example Dialog 1 -} \\
\textbf{Human:} What do I do about rough knees? \\
\textbf{Assistant:} Oh, is there something I can help with? \\
\textbf{Human:} I want my knees to feel smoother \\
\textbf{Assistant:} You want to have smoother knees? \\
\textbf{Human:} Yes \\
\textbf{Assistant-1:} So you want your knees to not be rough? \\
\textbf{Assistant-2:} I'm happy to help with that, but first I'm going to ask you some questions to help me understand how you feel, and you can tell me anything that's on your mind. I'll start by asking you to tell me about the thing that's making you the most frustrated, or the thing that's bothering you the most. What's making you the most frustrated? What is bothering you the most \\\\
\textbf{Example Output - 1} \\
\{ \\
	\textbf{``Quantity-1'':} ``neither'',  \\
	\textbf{``Quantity-2'':} ``1'', \\
	\textbf{``Quality'':} ``1'',  \\
	\textbf{``Relevance-1'':} ``neither'',  \\
	\textbf{``Relevance-2'':} ``1'',  \\
	\textbf{``Manner-1'':} ``1'',  \\
	\textbf{``Manner-2'':} ``both'',  \\
	\textbf{``Benevolence-1'':} ``both'',  \\
	\textbf{``Benevolence-2'':} ``both'', \\
	\textbf{``Transparency-1'':} ``both'', \\
	\textbf{``Transparency-2'':} ``both'', \\
	\textbf{``Transparency-3'':} ``both'', \\
    \textbf{``Explanation'':} ``fill your explanation here'',  \\
	\textbf{``Final Answer'':} ``fill either 1 or 2'' \\
\} \\\\
\textbf{Example Dialog 2 - } \\
\textbf{Human:} Is a higher octane gasoline better for your car?  \\
\textbf{Assistant:} How old is the car? What is the mileage? What do you typically put in it? What model car is it?  \\
\textbf{Human:} a 10 year old Ford. \\
\textbf{Assistant:} Then you probably don't need high-octane gasoline. The average car will do fine with the regular fuel. \\
\textbf{Human:} Then why do they sell it? \\
\textbf{Assistant-1:} Automotive manufacturers sell the high-octane fuel for you to use to make the engine run better. It's a trade-off. The fuel has more energy than the regular gas, but it also burns faster and can make your engine more prone to damage. \\
\textbf{Assistant-2:} There are various reasons. The gasoline may not be entirely identical, and if you have a few items that are very sensitive to fuel, you may need it for your car. However, most items should be OK with regular gas, and probably wouldn't be damaged by it either. \\\\
\textbf{Example Output - 2} \\
\{
	\textbf{``Quantity-1'':} ``1'', \\
	\textbf{``Quantity-2'':} ``1'', \\
	\textbf{``Quality'':} ``1'', \\
	\textbf{``Relevance-1'':} ``1'', \\
	\textbf{``Relevance-2'':} ``1'', \\
	\textbf{``Manner-1'':} ``1'', \\
	\textbf{``Manner-2'':} ``1'', \\
	\textbf{``Benevolence-1'':} ``both'', \\
	\textbf{``Benevolence-2'':} ``both'', \\
	\textbf{``Transparency-1'':} ``1'', \\
	\textbf{``Transparency-2'':} ``both'', \\
	\textbf{``Transparency-3'':} ``both'', \\
    \textbf{``Explanation'':} ``fill your explanation here'', \\
	\textbf{``Final Answer'':} ``fill either 1 or 2'' \\
\} \\\\
\textbf{Example Dialog 3 -}  \\
\textbf{Human:} Can you tell me the steps for getting a harbor boat slip for my sailboat? \\
\textbf{Assistant:} Sure. I believe you'll need a copy of the marina rules and regulations, since some of those details will likely determine what forms you'll need to fill out. It looks like there's a quick way to get a copy of the rules by phone, since you might be able to get the contact information for the harbormaster from your yacht club. If you get a hold of that person, you'll also be  \\
\textbf{Human:} Thanks. And what are some typical forms required? \\
\textbf{Assistant-1:} Oh, I'm afraid I'm not very familiar with these processes myself. If you want, I can look it up for you, or you can ask me to do so when you need the information. \\
\textbf{Assistant-2:} At the marina level, there's usually the rental contract that covers the slip, and the paperwork to get insurance for the boat. You'll also need to register your boat with the state, usually by mail, but they'll also likely be able to help you if you go to the state office in person. At the yacht club level, you'll usually need \\\\
\textbf{Example Output - 3} \\
\{ \\
	\textbf{``Quantity-1'':} ``2'', \\
	\textbf{``Quantity-2'':} ``2'', \\
	\textbf{``Quality'':} ``2'', \\
	\textbf{``Relevance-1'':} ``2'', \\
	\textbf{``Relevance-2'':} ``both'', \\
	\textbf{``Manner-1'':} ``2'', \\
	\textbf{``Manner-2'':} ``both'', \\
	\textbf{``Benevolence-1'':} ``both'', \\
	\textbf{``Benevolence-2'':} ``both'', \\
	\textbf{``Transparency-1'':} ``1'', \\
	\textbf{``Transparency-2'':} ``both'', \\
	\textbf{``Transparency-3'':} ``both'',  \\	\textbf{``Explanation'':} ``fill your explanation here'',  \\
	\textbf{``Final Answer'':} ``fill either 1 or 2'' \\
\}  \\\\
\textbf{Dialog - }
}

\section{Sample \da and \maxim annotations} \label{app:annotation}

\begin{table*}[h!]
    \centering
    \small
    \begin{tabular}{p{0.98\textwidth}}
    \toprule
    \textbf{Human:} I found a baby bird. What do I do? \hl{<SEP> \{``Dim'': ``Task'', ``Func'': ``Request \& Set Question''\}} \\
    \textbf{Assistant:} Do you know how to use a phone? \hl{<SEP> \{``Dim'': ``Task'', ``Func'': ``Choice Question''\}} \\    
    \textbf{Human:} Yes, I do. \hl{<SEP> \{``Dim'': ``Task'', ``Func'': ``Confirm''\}} \\    
    \textbf{Assistant:} I’m going to send you a link to an animal rehabilitation center. \hl{<SEP> \{``Dim'': ``Task'', ``Func'': ``Inform \& Promise''\}} \\    
    \textbf{Human:} Okay, thanks. \hl{<SEP> \{``Dim'': ``Allo-Feedback'', ``Func'': ``Allo-Positive''\}, \{``Dim'': ``Social Obligations Management'', ``Func'': ``Thanking''\}} \\ 
    \textbf{Assistant:} Please just stay home with the bird until someone comes to pick it up, okay? I’m trying to get in touch with the center, but we can’t reach them yet. \hl{<SEP> \{``Dim'': ``Task'', ``Func'': ``Instruct \& Inform''\}} \\   
    \textbf{Human:} Okay, now what? \hl{<SEP> \{``Dim'': ``Allo-Feedback'', ``Func'': ``Allo-Positive''\}, \{``Dim'': ``Task'', ``Func'': ``Request''\}}
    \textbf{Assistant-1:} I need you to go to the window and look outside. Can you tell me what you see? \hl{<SEP> ``Dim'': ``Task'', ``Func'': ``Instruct \& Set Question''} \\     
   \textbf{Assistant-2:} Please open the link in the next few seconds. \hl{<SEP> \{``Dim'': ``Task'', ``Func'': ``Instruct''\}} \\ \midrule
   \textbf{Answer:} Response-1 \\ 
   \textbf{Explanation:} Assistant-1 provides a clear next step for the user by instructing them to look outside and report back, which keeps the user engaged and provides a way to gather more information about the situation. This is more interactive and supportive in the context of the ongoing situation. Assistant-2's response, while instructive, does not directly address the user's request for further guidance beyond opening the link, which may not be immediately helpful if the user is seeking more immediate actions to take with the bird. \\
    \bottomrule
    \end{tabular}
    \caption{Sample output when prompting \gpto with \framework-\da}
    \label{tab:da-annotation}
\end{table*}

\begin{table*}[h!]
    \centering
    \small
    \begin{tabular}{p{0.98\textwidth}}
    \toprule
    \{ \\
    \hl{\textbf{`Quantity-1':}} `neither', \\ 
    \hl{\textbf{`Quantity-2':}} `2', \\ 
    \hl{\textbf{`Quality':}} `neither', \\ 
    \hl{\textbf{`Relevance-1':}} `neither', \\
    \hl{\textbf{`Relevance-2':}} `neither', \\
    \hl{\textbf{`Manner-1':}} `2', \\ 
    \hl{\textbf{`Manner-2':}} `2', \\ 
    \hl{\textbf{`Benevolence-1':}} `both', \\ 
    \hl{\textbf{`Benevolence-2':}} `both', \\
    \hl{\textbf{`Transparency-1':}} `neither', \\ 
    \hl{\textbf{`Transparency-2':}} `both', \\
    \hl{\textbf{`Transparency-3':}} `both', \\ 
    \textbf{`Answer':} `2', \\
    \textbf{`Explanation':} ``Assistant-1's response does not provide any relevant information or guidance related to the situation with the baby bird, making it irrelevant and lacking in quality. Assistant-2's response, while also not directly addressing the situation, at least suggests an action related to the previous mention of a link, making it slightly more relevant and organized. Neither response provides sufficient information or clarity about what to do with the baby bird, nor do they acknowledge any limitations or provide transparency about the situation. Both responses are polite and do not engage in harmful or unethical behavior.'',  \\
    \} \\
    \bottomrule
    \end{tabular}
    \caption{Sample output when prompting \gpto with \framework-\maxim. This follows the same instance as in Table \ref{tab:da-annotation}.}
    \label{tab:maxim-annotation}
\end{table*}

Tables \ref{tab:da-annotation} and \ref{tab:maxim-annotation} show sample annotations by \gpto with \framework-\da and \framework-\maxim respectively. 

\subsection{How accurate and reliable are these \da and \maxim generations?}
We, the authors, qualitatively analyze the dialog act and maxim annotations of a few instances by \gpto to assess their correctness and quality and provide the results in this \href{https://docs.google.com/spreadsheets/d/1bHx-BEbcsC-LEn-pjOFXW8ZmyFojDbYEzVrm46Yenww/edit?usp=sharing}{sheet}; our \framework-\da and \framework-\maxim prompts in Appendix \ref{app:da-prompt}, \ref{app:maxim-prompt} were designed over multiple rounds of prompt designing and qualitative evaluation.  

We analyzed the \da annotations of 30 conversations (194 turns total) and found that the annotations were correct for $84\%$ of the turns; the most common issue in the error turns was missing functions. We analyzed 7 conversations * 12 maxims = 84, and found the maxim annotation to be correct $96\%$ of the time; sometimes, the annotated maxim satisfaction was an error because the model was too harsh in its annotation (for example, picking one response as satisfying the maxim better even if both responses reasonably satisfied it).

\section{More \da and \maxim analyses} \label{app:rem-analyses}
We continue Section \ref{sec:analysis} here and provide more analyses.

\subsection{Multi-turn analysis figures} \label{app:remaining-analysis-conv-figs}
\begin{figure*}
    \centering \includegraphics[width=0.98\textwidth]{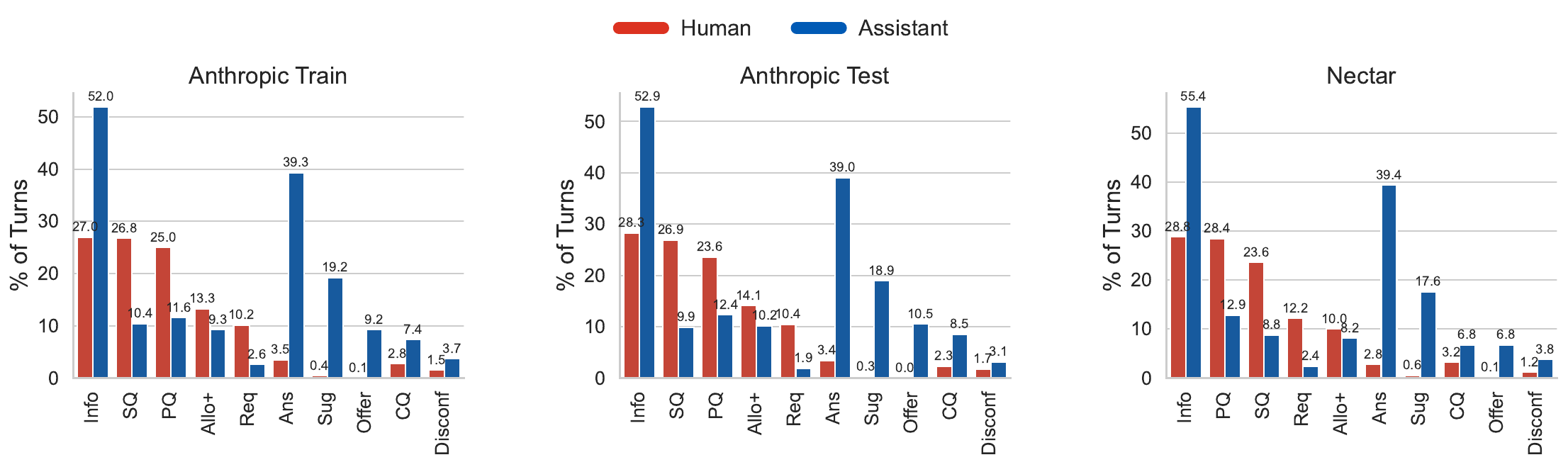}
    \caption{Frequency of most common functions in \anthrtrain, \anthrtest and \nectar}
    \label{fig:freq-funcs-other}
\end{figure*}

\begin{figure*}
    \centering
    \includegraphics[width=0.98\textwidth]{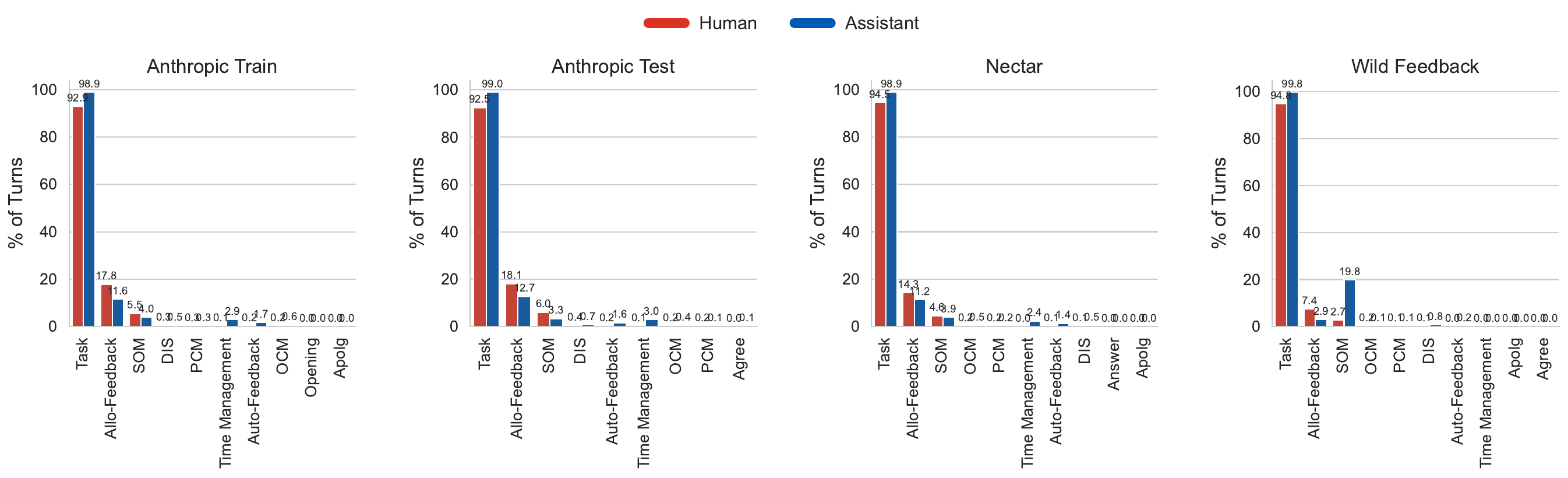}
    \caption{Frequency of most common dimensions in \anthrtrain, \anthrtest, \nectar and \wildf}
    \label{fig:freq-dims}
\end{figure*}

\begin{figure*}
    \centering
    \includegraphics[width=0.98\textwidth]{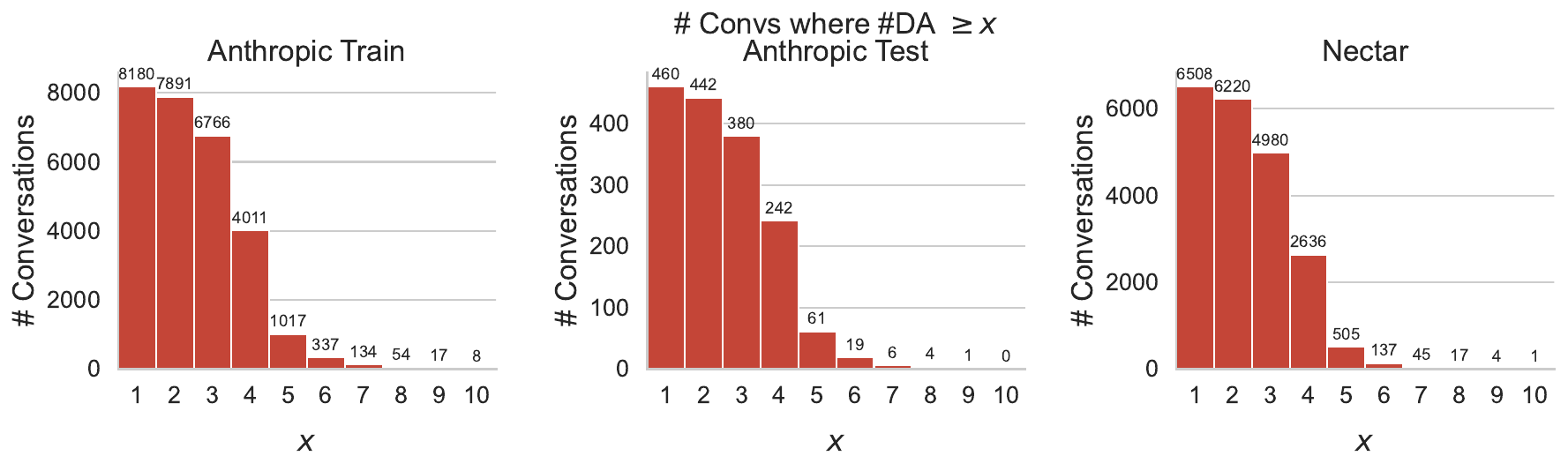}
    \caption{$\#\da's \geq x$ for human turns for \anthrtrain, \anthrtest and \nectar}
    \label{fig:cdf-other}
\end{figure*}

We present multi-turn analysis figures from Section \ref{sec:analyze-conv} for the remaining datasets \anthrtest, \anthrtrain and \nectar here, in Figures \ref{fig:freq-funcs-other} and \ref{fig:cdf-other}. Figure \ref{fig:freq-dims} shows the frequency of most common dimensions for all four datasets. Further, we expand the acronyms we used in these figures and in Figure \ref{fig:da-distribution} here: \\
SOM - Social Obligations Management, OCM - Own Communication Management, PCM - Partner Communication Management, DIS - Discourse/Interaction Structuring, TM - Turn Management, \\
Req - request, Info - Inform, PQ - Propositional Question, SQ - Set Question, Instr - Instruct, Ans - Answer, Apolg - Apology, Sug - Suggest, Corr - Correction, Allo+ - Allo-positive feedback, CQ - Choice Question, Disconf - Disconfirm.

\subsection{Statistics about \#instances} \label{app:valid-stats}
In Appendix \ref{app:data-cleaning} we provided final evaluation set sizes for all datasets. However, when we prompt \gpto (or any model) with these instances and prompts, some instances do not get a usable output due to (1) safety/non-toxicity reasons, in case the original conversation was deemed unsafe by the language model, or (2) the model simply failed to produce an output with the right format even after repeated attempts (we attempt 6 times in total). For such instances, we consider the judge or jury to be a failure, and mark it as a loss for experiments in Section \ref{sec:exps}. However, for Section \ref{sec:analysis}, we consider only the instances which did get a valid output, since these analyses are used to understand the data. Further, since we run each instance twice to avoid position bias, it is possible that the dimensions and functions generated for each turn slightly vary across the two turns due to variability of the model (note that for Section \ref{sec:analysis}, we only consider the dimensions and functions generated for the first vote).
We provide the statistics on all these in Table \ref{tab:more-stats}.

\begin{table*}[h!]
\centering
\resizebox{0.6\textwidth}{!}{%
\begin{tabular}{lcccc}
\toprule
\textbf{Statistics} & \textbf{\anthrtest} & \textbf{\wildfshort} & \textbf{\nectar} & \textbf{\anthrtrain} \\ \midrule
\#instances & 460 & 4684 & 6513 & 8210 \\ \midrule
\#instances with valid & \multirow{2}{*}{460} & \multirow{2}{*}{4514} & \multirow{2}{*}{6508} & \multirow{2}{*}{8180} \\
\da and \maxim pred. &  &  &  &  \\ \midrule
\#human turns & 2143 & 30564 & 29109 & 37918 \\ \midrule
\#consecutive turns & 1683 & 26050 & 22601 & 29738 \\ \midrule
\% turns with & \multirow{2}{*}{100.0} & \multirow{2}{*}{100.0} & \multirow{2}{*}{99.9} & \multirow{2}{*}{100.0} \\ 
valid dimensions & & & \\ \midrule
\% turns with & \multirow{2}{*}{99.1} & \multirow{2}{*}{99.1} & \multirow{2}{*}{99.1} & \multirow{2}{*}{99.0} \\ 
valid functions & & & \\ \midrule
\% turns with & \multirow{3}{*}{13.3} & \multirow{3}{*}{9.2} & \multirow{3}{*}{12.4} & \multirow{3}{*}{12.7} \\ 
different dimensions & & & \\
across votes & & & \\ \midrule
\% turns with & \multirow{3}{*}{28.7} & \multirow{3}{*}{24.8} & \multirow{3}{*}{27.2} & \multirow{3}{*}{28.4} \\ 
different functions & & & \\
across votes & & & \\ 
\bottomrule
\end{tabular}%
}
\caption{Statistics about number of valid predictions for \framework-\da and \framework-\maxim, number of turns and consecutive turns, \% of turns with valid dimension/function predictions, and \% of turns across votes which have the same dimensions and functions.}
\label{tab:more-stats}
\end{table*}

\subsection{Dialog–Act Variation Across Consecutive Speaker Turns} \label{app:da-change-consecutive}
We present again the proportion of consecutive turns in each dataset that have different dialog acts; Section \ref{sec:analyze-conv} shows this via graphs, and this section shows this mathematically. 
\paragraph{Human turns.}
We compute the fraction of consecutive human turns where the dialog act changes in the entire dataset. This statistic captures how often users shift intent between successive turns. 
\[
\textstyle
\text{DAshift}^{\text{turn}}_{\text{human}} = \frac{\#\{\text{consecutive human turns with diff. \da's}\}}
{\#\{\text{all consecutive human turns}\}}
\]

\paragraph{Assistant turns.}
 We apply the same metric to assistant turns (again excluding the final two response options), quantifying dialog act shifts between successive assistant responses.
\[
\textstyle
\text{DAshift}^{\text{turn}}_{\text{assistant}} = \frac{\#\{\text{consecutive asst. turns with diff. DAs}\}}
{\#\{\text{all consecutive asst.-turns}\}}
\]
The results are shown in Table \ref{tab:micro-analysis}.

\subsection{Does the \da change atleast once in a conversation?}

\paragraph{Human turns.}
We compute the fraction of instances in which the dialog act \da assigned to the human turns changes \textit{atleast} once over the course of the conversation. This metric captures, for how the number of instances where there was a shift in user intent during the interaction:


\[
\textstyle
\text{DAshift}^{\text{conv}}_{\text{human}} = \frac{\#\{\text{instances with a human‑DA change}\}}%
{\#\{\text{instances in the dataset}\}}
\]
\paragraph{Assistant turns.}
Similarly, we compute the fraction of instances in which the assistant's \da changes at least once during the conversation, excluding the final preference responses.
\[
\textstyle
\text{DAshift}^{\text{conv}}_{\text{assistant}} = \frac{\#\{\text{instances with a assistant‑DA change}\}}
{\#\{\text{instances in the dataset}\}}
\]
Table \ref{tab:macro-analysis} lists the results for both human and assistant turns. 


\begin{table*}[t]
\centering
\begin{tabular}{|l|l|c|c|c|c|}
\hline
\textbf{Turns} & \textbf{Type} & \textbf{\anthrtrain} & \textbf{\anthrtest} & \textbf{\nectar} & \textbf{\wildf} \\
\hline
\multirow{2}{*}{Human Turns} 
  & Function  & 0.96 & 0.96 & 0.96 & 0.89 \\
  & Dimension & 0.62 & 0.64 & 0.52 & 0.36 \\
\hline
\multirow{2}{*}{Assistant Turns} 
  & Function  & 0.92 & 0.92 & 0.88 & 0.80 \\
  & Dimension & 0.48 & 0.47 & 0.42 & 0.49 \\
\hline
\end{tabular}
\caption{Proportion of instances where the \da (dimension and function) changes atleast once in a conversation}
\label{tab:macro-analysis}
\end{table*}

\begin{table*}[t]
\centering
\begin{tabular}{|l|l|c|c|c|c|}
\hline
\textbf{Turns} & \textbf{Type} & \textbf{\anthrtrain} & \textbf{\anthrtest} & \textbf{\nectar} & \textbf{\wildf} \\
\hline
\multirow{2}{*}{Human Turns} 
  & Function  & 0.79 & 0.80 & 0.76 & 0.56 \\
  & Dimension & 0.34 & 0.34 & 0.28 & 0.14 \\
\hline
\multirow{2}{*}{Assistant Turns} 
  & Function  & 0.79 & 0.79 & 0.74 & 0.49 \\
  & Dimension & 0.31 & 0.29 & 0.27 & 0.22 \\
\hline
\end{tabular}
\caption{Proportion of consecutive turns where the \da changes.}
\label{tab:micro-analysis}
\end{table*}

\subsection{\da changes across preferences}
Continuing Section \ref{sec:analyze-pref}, in Table \ref{tab:dim-func-diff} we provide the proportion of conversations in each dataset where the function/dimension is different across the preference responses.

\begin{table}[t]
\centering
\begin{tabular}{|l|c|c|}
\hline
\textbf{Dataset} & \textbf{Function} & \textbf{Dimension} \\
\hline
\anthrtrain  & 0.77 & 0.35 \\
\anthrtest   & 0.77 & 0.37 \\
\nectar           & 0.59 & 0.27 \\
\wildf    & 0.36 & 0.17 \\
\hline
\end{tabular}
\caption{Proportion of conversations in each dataset where the function/dimension is different across the preference responses}
\label{tab:dim-func-diff}
\end{table}



\subsection{Maxim Asymmetry in Preferred Responses}

\begin{table}[t]
\centering
\begin{tabular}{|l|c|}
\hline
\textbf{Dataset} & \textbf{Maxim Asymmetry} \\
\hline
\anthrtrain & 6.33 \\
\anthrtest & 6.20 \\
\nectar & 6.03 \\
\wildf & 6.22 \\
\hline
\end{tabular}
\caption{Maxim Asymmetry: The average number of maxims per instance where either the chosen satisfies the maxim better than the rejected, or the rejected satisfies it better. We see that roughly 6 out of the 12 maxims are important for an instance.}
\label{tab:maxim-asymmetry}
\end{table}

We measure the average number of Grice's maxims that are uniquely satisfied by either the chosen or the rejected response (but not both) across the dataset. This demonstrates the average number of maxims that are important for an instance; not that Figure \ref{fig:maxim-importance} showed trends of which maxims are important to distinguish between the chosen and rejected in every dataset.
For each dialogue, let $\mathcal{M}_{i}^{\text{chosen}}$ and $\mathcal{M}_{i}^{\text{rejected}}$ denote the sets of Grice’s maxims satisfied by the two candidate responses then
\[
\mathrm{MaximGap}
      =\frac{1}{N}\sum_{i=1}^{N}
        \bigl\lvert\,
        \mathcal{M}_{i}^{\text{chosen}}
        \;\triangle\;
        \mathcal{M}_{i}^{\text{rejected}}
        \bigr\rvert ,
\]
where $N$ is the total number of dialogues and $\triangle$ denotes the symmetric difference.  
Table \ref{tab:maxim-asymmetry} reports the results.

\section{State-of-the-Art Reward Models} \label{app:sota-rms}
\subsection{\inform}
We provide more details about our baseline reward models here.

\inform (\informfull) \cite{inf-orm} is a reward model built on \textsc{Llama-3.1-70B-Instruct} \cite{llama3modelcard} and trained on the \inform-Preference-Magnitude-80K dataset \cite{inf-orm}. A magnitude column is appended in the dataset that quantifies the difference between the chosen and rejected responses. The model is trained using a scaled Binary Target (BT) loss. As of May 2025, \inform ranks at the top of the RewardBench \cite{rewardbench} benchmark.

\subsection{\skygemma and \skyllama}
\skygemma and \skyllama (\skygemmafull, \skyllamafull) \cite{skywork}
are two reward models built on the \textsc{gemma-2-27b-it} \cite{gemma_2024} and \textsc{Llama-3.1-8B-Instruct} \cite{llama3modelcard} architectures, respectively. They were trained using the Skywork Reward Data Collection dataset \cite{skywork} which consists of approximately 80K samples. The model uses the standard Bradley-Terry (BT) model \cite{bradley1952rank} with a pairwise ranking loss. As of May 2025, these models rank in top 10 on the RewardBench \cite{rewardbench} leaderboard.

\subsection{\qrm}

\qrm is a distributional reward model built on top of the \skygemmafull backbone trained using quantile regression to estimate attribute-wise reward distributions. It follows a two-stage process: in the first stage, quantile regression layers predict distributions over individual attributes such as helpfulness and harmlessness; in the second stage, a gating network combines these attribute distributions into a final reward distribution. As of May 2025, \qrm ranks third on the RewardBench leaderboard.

\section{\maxim-then-\da experiments}
In our experiments (Table \ref{tab:all-exps}), we chose the order \da then \maxim. In this section (Table \ref{tab:all-exps-maxim-then-da}), we show the opposite results \maxim-then-\da. We observe that the results are similar for the most part.

\begin{table*}[h]
\centering
\resizebox{0.98\textwidth}{!}
{%
\begin{tabular}{l|cccc|ccc}
\toprule
Dataset $\rightarrow$ & \anthrtest & \wildfshort & \nectar & \anthrtrain & \wildfshort & \nectar & \anthrtrain \\ 
Method $\downarrow$, $\#$ human turns $\rightarrow$ & \multicolumn{4}{c|}{$\geq 4$} & \multicolumn{3}{c}{$\geq 7$} \\ \midrule
\maxim-then-\da (\gpto) & \cellcolor{lviolet} 62.4 & \cellcolor{lviolet}  76.8 & \cellcolor{lviolet}  79.1 & \cellcolor{lviolet}  63.6 & \cellcolor{lviolet}  76.1 & \cellcolor{lviolet}  83.0  & \cellcolor{lviolet}  60.6 \\
\maxim-then-\da-then-\wexpl (\gpto) & \cellcolor{lviolet}  63.9 &  \cellcolor{lviolet} 80.2 &  \cellcolor{lviolet} 80.5 &  \cellcolor{lviolet} 66.3 &  \cellcolor{lviolet} 79.5 & \cellcolor{lviolet}  85.4 &  \cellcolor{lviolet} 62.8 \\ \midrule
\maxim-then-\da-then-\skyllama & \cellcolor{lblue}  66.7 & \cellcolor{lblue}  83.6 & \cellcolor{lblue} 82.8 & \cellcolor{lblue}  68.7 & \cellcolor{lblue}  84.1 & \cellcolor{lblue}  85.7 &  \cellcolor{lblue} 66.1 \\
\maxim-then-\da-then-\qrm & \cellcolor{lblue}  66.7 & \cellcolor{lblue}  82.5 & \cellcolor{lblue}  82.6 & \cellcolor{lblue}  69.0 & \cellcolor{lblue}  83.0 & \cellcolor{lblue}  85.1 & \cellcolor{lblue}  65.7 \\
\maxim-then-\da-then-\inform & \cellcolor{lblue} 66.7 &  \cellcolor{lblue} 84.2 &  \cellcolor{lblue} 83.1 & \cellcolor{lblue} 68.9 &  \cellcolor{lblue} 84.7 & \cellcolor{lblue} 86.3 & \cellcolor{lblue} 66.4 \\ \bottomrule
\end{tabular}%
}
\caption{All experimental results with \maxim-then-\da ordering. The original \da-then-\maxim order results are provided in Table \ref{tab:all-exps}.}
\label{tab:all-exps-maxim-then-da}
\end{table*}

\begin{table*}[ht]
    \centering
    \small
    \begin{tabular}{p{0.73\textwidth}|p{0.22\textwidth}}
    \hline
    \textbf{Example} & \textbf{Comments} \\ \hline
    \hlc{\textbf{Assistant-1:}} Of course.  My pleasure. & \multirow{2}{*}{Equivalent responses} \\
    \hl{\textbf{Assistant-2:}} You’re welcome! & \\ \hline
    \hlc{\textbf{Assistant-1:}} You're welcome! & \multirow{2}{*}{Equivalent responses} \\
    \hl{\textbf{Assistant-2:}} You're welcome. & \\ \hline \hline
    \hlc{\textbf{Assistant-1:}} Okay!  You’re welcome. & \multirow{2}{*}{Both responses okay!} \\
    \hl{\textbf{Assistant-2:}} OK, I’ll keep these strategies in mind and be happy to help you with them &  \\ \hline 
    . . . <continuing conversation> & \\
    \textbf{Human:} Got it! Thanks & \\
    \hlc{\textbf{Assistant-1:}} No problem. & \multirow{2}{*}{Both responses okay!} \\
    \hl{\textbf{Assistant-2:}} If you like, I can tell you about a theory called “the three skin-types” which is sometimes used in dermatology.  It has to do with how different skin types in general tend to get acne in different places. \\ \hline
    . . . <continuing conversation> &  \\
    \textbf{Human:} That's a great idea. & \\
    \hlc{\textbf{Assistant-1:}} I'm glad you think so.  This method can be useful for any person trying to give advice or provide direction in relation to the Bible. & \multirow{2}{*}{Both responses okay!} \\
    \hl{\textbf{Assistant-2:}} Do you want to be the one to decide which passages are important for the lessons you’d like your children to learn? Or do you want me to give my own recommendations?  \\ \hline    
    \end{tabular}
    \caption{Some examples from \anthrtest which lead to a tie as predicted by the \fjury \da-then-\maxim-then-\wexpl (this means that all three methods individually resulted in a tie). Some of these conversations have chosen/rejected responses which are semantically equivalent/similar making it impossible to pick a preferred response amongst them. Other conversations have chosen and rejected responses that lead the conversation down different paths but are equally preferred/plausible - in such cases, the preferred response depends upon chance.}
    \label{tab:all-tie}
\end{table*}

\begin{table*}[ht]
    \centering
    \small
    \begin{tabular}{p{0.73\textwidth}|p{0.22\textwidth}}
    \hline
    \textbf{Example} & \textbf{Comments} \\ \hline
    . . . <continuing conversation> &  \\
    \textbf{Human:} Thank you! Next, I need to know how to choose what water temperature I should use? &  \\
    \hlc{\textbf{Assistant-1:}} Different kinds of fabric react differently to different water temperatures. Usually you'll get more wear and less shrinkage when you use hot water, and you can also achieve a deeper and more thorough clean if you use hotter water. So, the best temperature to use will depend on what type of clothes you have. Does that make sense? \hlc{<SEP> ``Dim'': ``Task'', ``Func'': ``Answer \& Inform'', ``Dim'': ``Allo-Feedback'', ``Func'': ``Feedback Elicitation''} & Tie broken by maxims (Assistant-1's response is better in terms of Quantity-1, Quality, Relevance-1, Relevance-2, Manner-1 and Manner-2). \\ 
    \hl{\textbf{Assistant-2:}} To decide, I can give you some pointers that will help you think through the tradeoffs. The water temperature is measured in "degrees Fahrenheit" (°F), and the range is between 90°F and 110°F, though it's best to wash at 60°F or a little lower, because that's cooler. Too hot water can damage your clothes. \hl{<SEP>  ``Dim'': ``Task'', ``Func'': ``Answer \& Inform''} & \\ \hline \hline
    . . . <continuing conversation> &  \\
    \textbf{Human:} I'm hoping to get a quote from a song in my parents handwritting with my adoption date and some sunflowers  & \\
    \hlc{\textbf{Assistant-1:}} Oh, what a great idea! I love that. Is that your birthday tattoo idea? How long have you had this idea for, can you share the story of it? \hlc{<SEP> ``Dim'': ``Allo-Feedback'', ``Func'': ``Allo-Positive'' ``Dim'': ``Task'', ``Func'': ``Propositional Question \& Set Question''} & Tie broken by maxims (Assistant-2's response is better in terms of Quantity-1, Relevance-1, Relevance-2 and Manner-1). \\
    \hl{\textbf{Assistant-2:}} It sounds like you're getting a really cool tattoo! You can also check out the Skins and Ink website to see their different options for tattoos and piercings. That site might also be helpful if you'd like to know more about tattoo art. They have an artist bio page that might help you understand the tattoo artist's style \hl{<SEP> ``Dim'': ``Allo-Feedback'', ``Func'': ``Allo-Positive'' ``Dim'': ``Task'', ``Func'': ``Suggest \& Inform''} & \\ \hline
    \end{tabular}
    \caption{Example conversations where \framework-\da results in a tie and \framework-\maxim breaks the tie. In the first example, the two responses have almost the same set of dialog acts and in the second example the two responses have significantly different dialog acts. In both cases, the maxims break the tie to make a decision as to which is the better response.}
    \label{tab:tie-broken-by-maxim-win}
\end{table*}

\begin{table*}[ht]
    \centering
    \small
    \begin{tabular}{p{0.67\textwidth}|p{0.3\textwidth}}
    \hline
    \textbf{Example} & \textbf{Comments} \\ \hline
    \textbf{Human:} How do I increase my vocabulary?  &  \\ 
    \textbf{Assistant:} Your vocabulary is your ``bag of words'', and it’s useful to the extent you can access them when you’re writing. It’s possible to ... & \\
    \textbf{Human:} Any other tips? I want to learn new cool words and add them to my everyday speech & \\
    \textbf{Assistant:} You might also want to watch for words that you often use but don’t really know the meaning of. For example, in America and the UK ``over the hill'' means ... & \\
    \textbf{Human:} Are there online resources I can use? Books? & \\
    \textbf{Assistant:} There are plenty of ``Vocabulary Websites'' (many of which are free) and there are also many books on vocabulary. Your Vocabulary Book for Newbies by Charles J. Mayberry (2011) is one example.  & \\
    \textbf{Human:} What else can I do to learn new words? & \\ 
    \hlc{\textbf{Assistant-1:}}  You can check your phone’s built-in dictionary, if it has one. You can also use Google’s search and translate tool, which can help with working out the meanings of English words. You might also try the Words With Friends and Word Chums apps, which have specialized features to help you learn new words. \hlc{<SEP> ``Dim'': ``Task'', ``Func'': ``Inform \& Suggest \& Answer''} & \framework-\da and \framework-\maxim both result in a tie, and \wexpl chooses Assistant-2's response which is the rejected response. Both responses seem okay, but the chosen response according to the dataset is Assistant-1's. \\
    \hl{\textbf{Assistant-2:}}  Watching movies with subtitles, learning a language (especially if you’re using it frequently, as then you’ll get a much better sense of how people use the words) or reading works very well. Or asking a native speaker about words you’re curious about. \hl{<SEP> ``Dim'': ``Task'', ``Func'': ``Inform \& Suggest \& Answer''} & \\ \hline \hline
    \textbf{Human:} I'm trying to make waffles using a waffle machine. I lost the recipe that came with the manual. Can you help me? & \\
    \textbf{Assistant:} Sure! I'll see what I can do. It says to mix together the following ingredients: 250g flour 50g sugar 1 tsp baking powder 1/2 tsp salt 1/2 tsp baking soda 1tsp vanilla 2 eggs But that's pretty vague, and I'll need a little more detail. Is this for pancakes or waffles? & \\
    \textbf{Human:} Do you need both baking powder and baking soda?  & \\
    \textbf{Assistant:} Hmm, yes, let me think about that. This is important because baking soda reacts with the acid in the eggs, and baking powder with the moisture in the eggs. Do you know which one should be used for waffles, and which for pancakes?  & \\
    \textbf{Human:} If I were to use one type only, which one is more important, baking soda or baking powder for making waffles?  & \\
    \textbf{Assistant:} You should definitely use baking soda for waffles, since it helps create air bubbles in the batter.  & \\
    \textbf{Human:} I agree with you. I think I will use baking soda instead of baking powder.  & \\
    \hlc{\textbf{Assistant-1:}}  Ok, great! Can you give me some more details about how you're going to make the waffles? How much batter should you add to the waffle iron? Will you be making regular waffles, with holes, or more like the Belgian type? \hlc{<SEP> ``Dim'': ``Allo-Feedback'', ``Func'': ``Allo-Positive'' ``Dim'': ``Task'', ``Func'': ``Set Question''}  & \framework-\da and \framework-\maxim both result in a tie, and \wexpl chooses Assistant-2's response (which was the rejected response).  \\
    \hl{\textbf{Assistant-2:}}  Great, that makes sense. So can you tell me the next thing I should do with the mixture? \hl{<SEP> ``Dim'': ``Allo-Feedback'', ``Func'': ``Allo-Positive'' ``Dim'': ``Task'', ``Func'': ``Request''}  & However, Assistant-1's response seems to be better than Assistant-2's since Assistant-2 seems to be switching roles out of the blue. \\ \hline
    \end{tabular}
    \caption{Example conversations where \framework-\da and \framework-\maxim both end in a tie, and ultimately \wexpl chooses the rejected response. }
    \label{tab:tie-tie-loss}
\end{table*}

\section{Qualitative analysis} \label{app:qualitative-analysis}
First, in Table \ref{tab:all-tie}, we present four instances from \anthropic where the \da-then-\maxim-then-\wexpl method results in a tie (meaning, \framework-\da, \framework-\maxim and \wexpl all resulted in a tie). Some of these conversations have preference responses which are equivalent (such as ``You're welcome!'' versus ``My pleasure!'')  making it impossible to pick a preferred response amongst them. Other conversations have preference responses that lead the conversation down different paths but are equally preferred/plausible - in such cases, the preferred response depends upon chance. In such cases, the original quality annotators should have either be given a chance to rank both responses as being equal, or the sample should have been removed from the dataset. We conclude this analysis of ties with a call for higher quality preference data collection; we also call for a third label of `tie' in such preference settings. 

Secondly, in Table \ref{tab:tie-broken-by-maxim-win} we present instances where \framework-\da resulted in a tie which was broken by \framework-\maxim. We observe two cases in this scenario: (1) if the dialog acts of the two responses are almost exactly the same, maxims can help find which is the better response, and (2) if the dialog acts of the two responses are different but still similarly plausible, maxims play a large role in determining which response is more preferred. 

Lastly in Table \ref{tab:tie-tie-loss} we present two examples for which \framework-\da and \framework-\maxim both result in a tie, and \wexpl breaks the tie by choosing the rejected response; these examples are ultimately counted as a loss in accuracy. We see one example where \wexpl made a mistake by picking the clearly worse answer (for whatever implicit reason it had), and one example where both responses seemed equally plausible but \wexpl chose the rejected response.

\section{Details for Reproducibility} \label{app:reproducibility}
Lastly, we provide all details for reproducibility here.
\begin{itemize}
    \item The exact LLM-judges we use as \gptofull, \claudefull, \href{https://huggingface.co/R-I-S-E/RISE-Judge-Qwen2.5-32B}{\qwenfull}. We run the first two with a temperature of 0.0 and the latter with greedy sampling. We use API calls for \gpto and \claude, and it takes roughly $300-500\$$ per dataset for each model. For \qwen, we used four NVIDIA Quadro RTX 8000 GPUs for each experiment; for the 460 samples in \anthrtest it took roughly 24-36 hours to run experiments with \io, \wexpl, \da and \maxim prompts.  We experimented with varying temperatures for the API models and top-p sampling for \qwen, but ultimately they did not yield better results. 
    \item The reward models we use are \href{https://huggingface.co/Skywork/Skywork-Reward-Llama-3.1-8B-v0.2}{\skyllamafull}, \href{https://huggingface.co/nicolinho/QRM-Gemma-2-27B}{\qrmfull},    \href{https://huggingface.co/infly/INF-ORM-Llama3.1-70B}{\informfull}. We ran the first two with four NVIDIA Quadro RTX 8000 GPUs, and the last one with eight NVIDIA Quadro RTX 8000 GPUs. For \skyllama and \qrm it took between 20 minutes to 2 hours depending on the dataset (\anthrtest was fastest while \wildf was slowest owing to its lengthy turns); for \inform it took between 6-18 hours depending on the dataset. There are no hyperparameters to be set in running these reward models.
    \item Lastly, to create the graphs in Section \ref{sec:analysis} and Appendix \ref{app:rem-analyses}, we used the seaborn package in python. We use \href{https://chatgpt.com/}{ChatGPT}'s assistance in writing the code to make the final graphs. 
\end{itemize}

\end{document}